\documentclass[journal]{IEEEtran}
\ifCLASSINFOpdf
\else
\fi
\usepackage{url}

\usepackage{amsmath}
\usepackage{color}
\usepackage{comment}
\usepackage{graphicx}

\usepackage{subfig}

\usepackage{floatrow}

\usepackage[usenames,dvipsnames]{xcolor}
\usepackage{tcolorbox}

\begin{document}
%
\title{Building Extraction at Scale using Convolutional Neural Network: Mapping of the United States}
%
%
%
\author{\IEEEauthorblockA{Hsiuhan Lexie Yang,~\IEEEmembership{Member,~IEEE,}
        Jiangye Yuan,~\IEEEmembership{Member,~IEEE,}
        Dalton Lunga,~\IEEEmembership{Senior Member,~IEEE,}
        Melanie Laverdiere, Amy Rose, Budhendra Bhaduri}

\IEEEauthorblockA{Computing and Computational Sciences Directorate, 
Oak Ridge National Laboratory, U.S.}
%
%
\thanks{This manuscript has been authored by UT-Battelle, LLC under Contract No. DE-AC05-00OR22725 with the U.S. Department of Energy. The United States Government retains and the publisher, by accepting the article for publication, acknowledges that the United States Government retains a non-exclusive, paid-up, irrevocable, world-wide license to publish or reproduce the published form of this manuscript, or allow others to do so, for United States Government purposes. The Department of Energy will provide public access to these results of federally sponsored research in accordance with the DOE Public Access Plan.}

}

\markboth{Journal of \LaTeX\ Class Files,~Vol.~14, No.~8, August~2015}%
{Shell \MakeLowercase{\textit{et al.}}: Bare Demo of IEEEtran.cls for IEEE Journals}
%



\maketitle

\begin{abstract}
Establishing up-to-date large scale building maps is essential to understand urban dynamics, such as estimating population, urban planning and many other applications. Although many computer vision tasks has been successfully carried out with deep convolutional neural networks, there is a growing need to understand their large scale impact on building mapping with remote sensing imagery.

Taking advantage of the scalability of CNNs and using only few areas with the abundance of building footprints, for the first time we conduct a comparative analysis of four state-of-the-art CNNs for extracting building footprints across the entire continental United States. 
The four CNN architectures namely: branch-out CNN, fully convolutional neural network (FCN), conditional random field as recurrent neural network (CRFasRNN), and SegNet, support semantic pixel-wise labeling and focus on capturing textural information at multi-scale. We use 1-meter resolution aerial images from National Agriculture Imagery Program (NAIP) as the test-bed, and compare the extraction results across the four methods. In addition, we propose to combine signed-distance labels with SegNet, the preferred CNN architecture identified by our extensive evaluations, to advance building extraction results to instance level. We further demonstrate the usefulness of fusing additional near IR information into the building extraction framework. Large scale experimental evaluations are conducted and reported using metrics that include: precision, recall rate, intersection over union, and the number of buildings extracted. With the improved CNN model and no requirement of further post-processing, we have generated building maps for the United States with an average processing time less than one minute for an area of size $\sim56$ $km^2$. The quality of extracted buildings and processing time demonstrated the proposed CNN-based framework fits the need of building extraction at scale.
\end{abstract}


%
\IEEEpeerreviewmaketitle

\section{Introduction}
%
%
%
%

\IEEEPARstart{S}{ince} high resolution remote sensing imagery became more accessible and affordable, extracting buildings by leveraging the cost-effective and fast-updated remote sensing images has been of great practical interest. The established building maps are used to understand urban dynamics, such as estimating population and facilitating urban planning, and many other applications in socio-economics studies \cite{JensenCowen1999}. Although many works have been devoted to enabling automated building extraction \cite{Ok2013a,Ngo2017,Li2015,Kim1999,Hermosilla2011}, there remains a challenge to establishing a reliable building footprints database at scale based on remote sensing imagery. This challenge arises from certain unscalable assumptions or the limited building hypotheses those algorithms made in order to achieve satisfying results. The increasingly popular Volunteered Geographic Information (VGI) from OpenStreetMap \cite{HaklayWeber2008} might be considered as a source of obtaining large scale building maps, however, the inconsistent quality of VGI usually leads to more efforts to refine the data before further use. As a result, a generic, robust and automated scalable framework for generating large scale building maps is yet to be  achieved.

Recent developments in deep convolutional neural networks (CNN) provide an unique opportunity to achieve remarkable object detection performance in the computer vision society. Most recently, there has been a surging interest from the remote sensing community with several works investigating the application of deep CNN toward building mapping \cite{PaisitkriangkraiSherrahJanneyEtAl2016, Yuan2018, Bittner2017481, 7729471,Lunga2018}. Even though the works are notable, there remains a gap to perform a comprehensive study that leverages the power of current state-of-the-art CNNs in examining large scale automated building extraction with remote sensing imagery. 

In this paper, we seek a scalable and reliable building extraction framework that encompasses state-of-the-art CNNs. Our unique contributions are three-fold: 
\begin{itemize}

	\item In order to evaluate the applicability and generalization of CNN models, we investigate and conduct an extensive validation process on the performance of several state-of-the-art CNNs over several testing sites in the United States. Although these approaches are widely applied and validated within the computer vision community, we assess on their suitability including challenges and opportunities for efficient large scale building extraction.
	\item We further incorporate custom designed signed-distance labels for more precise building outline extraction, aiming to elevate the building extraction performance to instance level. The goal of our building extraction task is not only to detect building location, we also desire the completeness of the  precisely extracted individual buildings (spatial extent). We also propose a simple but effective fusing strategy to combine two CNN models trained with additional spectral bands while still leveraging the learned parameter values of a pre-trained model for initialization. As to be demonstrated in our comprehensive experiments, the fusing approach yields a desirable performance boost.
	
	\item Using a single optimal CNN model derived via the validation process, we generate the first seamless building maps for the contiguous United States with a GPU cluster. Upon performing quality check on the results, we identify the major sources of commission errors. We then further refined building extraction results with a minimal re-training process. The achieved results demonstrates the use of deep CNN for robust building extraction at scale.  The insights provided by this operational task are benefiting future similar large scale object detection works based on remote sensing imagery.
\end{itemize}

\section{Related Work}


\subsection{Building Extraction using Remote Sensing Imagery}
Building mapping plays a prominent role in urban planning and population modeling, helping decision makers understand human activity in socioeconomic \cite{Jensen1999} and many other geospatial applications. Being a cost-effective data resource, remote sensing data has been considered as the primary resource to generate building maps \cite{Ok2013a,Ngo2017,Li2015,Kim1999,Hermosilla2011,Cote2013}. The approaches include edge-based, model-based \cite{Ngo2017} or evolution/energy based methods \cite{Cote2013}. Although many techniques have been proposed, including combining different spectral information \cite{Ok2013a}, simplifying building hypotheses \cite{Kim1999} and other auxiliary constraints \cite{Ngo2017}, and have shown some successful results, a generic, robust and scalable solution is not yet available. The major barrier to achieve reliable building extraction at scale with existing methods is that prior information and assumptions are not generalizable over extended areas. For example, with traditional model based approaches, the complexities of buildings can be understood by those methods only when the models or the shapes of buildings are provided. To extract buildings at scale, it is nearly impossible to establish a building shape database encompassing the entire country, the United States for example. Furthermore, the computational complexity and the process to optimize parameters poses limitations on extracting buildings effectively and efficiently. For production scale work, a building extraction framework should be with minimal effort to tuning feature extractors while generalizing well over large extent areas.

\subsection{Deep CNN for Building Extraction}
In contrast to the hand-crafted features based on prior information, learning feature extractors via multi-layer convolutional neural network (CNN) demonstrates a more powerful generalization capability  \cite{Goodfellow2016}. The pioneer of CNN LeNet \cite{LeCunBottouBengioEtAl1998} was proposed to successfully recognize the digits in the classical dataset (DIGITS) in the late 1990s. However, the heavy computational requirement hindered further developments for advancing CNN performance until the idea of utilizing GPU for neural computation is proposed. Enabled by the hardware advances, recent works further indicate that deeper network architectures (i.e. with many layers) often yield better performance. This has been the backbone of the recent surging interest to apply deep CNNs to a variety of applications \cite{SimonyanZisserman2015,SzegedyLiuJiaEtAl2015,KarpathyTodericiShettyEtAl2014,OquabBottouLaptevEtAl2014,SainathKingsburySaonEtAl2015}.

For exploiting remote sensing data, CNN also offers us much needed insights to leverage the art of automating feature learning and to promote better generalizability while minimizing manual efforts to hand-craft features. For example, one of the early works in \cite{PaisitkriangkraiSherrahJanneyEtAl2016} showed that CNN provides outstanding results on the benchmark ISPRS data set with little human intervention.

Although the re-emerged interests in CNN lead to significant success especially in image classification and object detection for natural image applications, earlier CNNs could not be directly used for building extraction with overhead imagery. Several widely known CNNs such as VGGNet \cite{Simonyan2015}, and GoogLeNet \cite{SzegedyLiuJiaEtAl2015} were designed for image classification at patch level, which means that each image patch is assigned a single class label. These CNNs have also been exploited as preferred frameworks for extracting abstract features from remote sensing imagery for scene understanding. However, extracting buildings from remote sensing images requires semantic segmentation, also known as pixel-wise labeling, so that we can obtain complete building outlines. It is possible to directly apply patch-based methods to generate pixel-wise labeling, however, it suffers from redundant convolution operations \cite{ShelhamerLongDarrell2017}. 

Another major issue pertaining to use of patch-level image classification CNNs in semantic segmentation is that after a series of pooling operations following convolutions the resulting feature maps will be smaller than the original input image \cite{Goodfellow2016}. Having an pixel-wise labeled output with the same resolution as the original input image is critical for remote sensing imagery understanding. 

Many works have been dedicated to enabling pixel labeling by compensating the coarser feature maps after convolution and pooling operations in classification CNNs \cite{NoordPostma2017}. One of such pioneering work that leverages patch level based classification CNN for semantic segmentation is the Fully Convolutional Network (FCN) \cite{ShelhamerLongDarrell2017}, which has also been adapted by the remote sensing community \cite{Sherrah2016}. The core concept in FCN for dense labeling is to replace fully connected layers in the patch-based CNN with convolutional layers. FCN also proposes upsampling operations, which can also be cast as convolution operation. A similar technique of exploring fully convolutional architecture is also proposed in \cite{Maggiori2016} to improve the results of the patch-based CNNs. 

It is noted that although including other supporting features after learning could potentially boost the dense labeling performance \cite{PaisitkriangkraiSherrahJanneyEtAl2016}, we prefer the end-to-end training scheme as in \cite{Yuan2018,ShelhamerLongDarrell2017,Maggiori2016,VolpiTuia2017} without post-proccessing or additional refinement strategy. The end-to-end framework allows more repeatable and scalable evaluations, especially for large scale building mapping that we are interested in this paper.

\subsection{Pre-trained Models}
Similar to other supervised learners, deep CNN requires sufficient training samples to estimate significant amount of parameters and often encounters some challenges in training  \cite{ErhanManzagolBengioEtAl2009}. As a result, long training time is often mandatory to obtain a deep CNN that delivers satisfactory performance.  It is surprisingly concluded by many works that exploiting the transferibility of pre-trained CNN models can yield better results \cite{Penatti2015} with even shorter training time to fine-tune the adapted parameters for a different task or with different data \cite{Bittner2017481}. In \cite{TajbakhshShinGuruduEtAl2016,Gerrand2017}, various medical imaging modalities were tested including training from scratch and as well as finetuning of pre-trained models that were trained with natural RGB images. The comprehensive comparisons have shown that fine-tuning pre-trained models, even with considerably different medical images, provides not only better image classification performance and segmentation tasks, but also help achieve training convergence faster. In addition, recent work has demonstrated that patch-based convolutional neural networks trained by the ImageNet dataset can generate satisfactory results for overhead imagery analysis \cite{ChengMaZhouEtAl2016}.

\section{Deep CNN with Pre-Trained Models for Building Extraction}
In this section, we review state-of-the-art (SOA) pixel labeling CNN methods and their consideration for large scale building extraction work. We demonstrate that deep CNNs can benefit from the notion of pre-trained classification models and proceed to generate segmentation result at pixel level. Furthermore, we propose an approach that is devised to better delineate building outlines and explore an additional spectral band that cannot be directly incorporated into pre-trained models without further adapting and learning weights of the input layer. 

\subsection{Fully Convolutional Network}
Fully convolutional network (FCN) \cite{ShelhamerLongDarrell2017} is the first work that effectively converts classification deep CNNs for dense labeling. This key feature permits FCN to take advantage of pre-trained classification CNN model and to generate prediction maps that are the same size as the input images. FCN uses \textit{transposed convolution} (or called fractionally strided convolution \cite{RadfordMetzChintala2015}) to upsample those final feature maps to match the original input image. The strategy of incorporating feature maps from different scales, called skip layer, is also exploited in FCN. The feature maps from shallower and deeper layers are fused to make predictions concerning both local and global features, respectively. With many benchmark data sets, FCN has demonstrated impressive performance in accuracy as well as computational time. The similar upsampling strategy is investigated in building extraction with remote sensing data \cite{Yuan2018,VolpiTuia2017,Maggiori2016} to create the densely labeled classification map. In \cite{Yuan2018} and \cite{Maggiori2016}, they further highlight that combining feature maps from multiple scales to capture features at both coarse and fine levels is beneficial for building extraction. 

However, the results of FCN are lacking boundary details for small objects such as small buildings. Identifying accurate boundary for buildings is particularly important in the building extraction task, as we aim to establish a building footprint map that provides the outline of a building drawn along the exterior walls, with a description of the exact size, shape, and its location. All this information can only be extracted by well-delineated building boundaries.

Since FCN was proposed, many of its variants, such as including a plug-in conditional random field module \cite{ZhengJayasumanaRomera-ParedesEtAl2015} (CRFasRNN) and SegNet \cite{BadrinarayananKendallCipolla2017},  have been proposed to advance the performance of dense semantic labeling, especially for better capturing object boundaries. 
\subsection{Conditional Random Field as Recurrent Neural Networks}
Conditional random field (CRF) has been used in segmentation problem to enhance the accuracy of pixel-level labeling. This graphical-model based approach is able to refine coarse pixel level label predictions to produce sharp boundaries and fine-grained segmentation results. A similar idea of exploiting CRF has also been proposed in building extraction work \cite{Li2015}. Although CRF can be used as a post-processing step to boost the FCN performance to address the boundary issue, the disconnect between CNN training and using CRF in post processing leads to sub-optimal performance in both mechanisms. In \cite{ZhengJayasumanaRomera-ParedesEtAl2015}, the strength of CRF and CNN is combined into a unified framework CRFasRNN. In CRFasRNN, homogeneous pixels are grouped during CNN training, and the boundaries of the objects are gradually refined. More importantly, the CRF module is cast as a recurrent neural network and is part of optimization during the CNN training. Such an end-to-end framework is favorable when large scale data processing is needed.   

\subsection{SegNet}
SegNet \cite{BadrinarayananKendallCipolla2017} is also proposed to further address the imperfect boundary delineation issue observed in FCN and other similar multi-stage CNNs for semantic segmentation. In SegNet, the upsampling strategy is also exploited, but compared to FCN the upsampled feature maps take the indices of the max-pooled feature maps into account. The benefits provided by SegNet are two-fold for our goal: 1)The stored indices of max-pooled feature maps capture strong responses from the convoluted inputs and they usually correspond to location of the edges of objects, which are buildings in our case. 2) Even though the inference time tends to be a little longer in SegNet, the memory required in SegNet is significantly less than FCN \cite{BadrinarayananKendallCipolla2017} or CRFasRNN. As a result, we can process a larger area faster in parallel at the inference stage when large scale processing is needed. In addition, since the encoder part of SegNet is identical to those popular classification nets such as VGGNet \cite{SimonyanZisserman2015}, we are able to take advantage of the pre-trained models to initialize and train a model specifically for the building extraction task.

\section{Proposed Approach}
Intuitively, identifying buildings from imagery is a binary classification problem, where buildings are one class and non-buildings are the other class. Although the above CNNs are capable of providing pixel labeling segmentation, we consider our building extraction problem at another level - from class-aware segmentation to instance-aware pixel labeling segmentation \cite{silberman2014instance,WangBaiMattyusEtAl2016}. This is especially crucial for building extraction in urban area , densely populated area where small buildings are in close proximity. Without considering building extraction at instance level, even though those buildings might be able to be extracted by CNNs, these kind of results provide limited information for broader applications that require precise area of buildings, number of buildings, etc \cite{JensenCowen1999}. 

\begin{figure} 
    \centering
  \subfloat[Binary label]{%
       \includegraphics[width=4cm,height=4cm ]{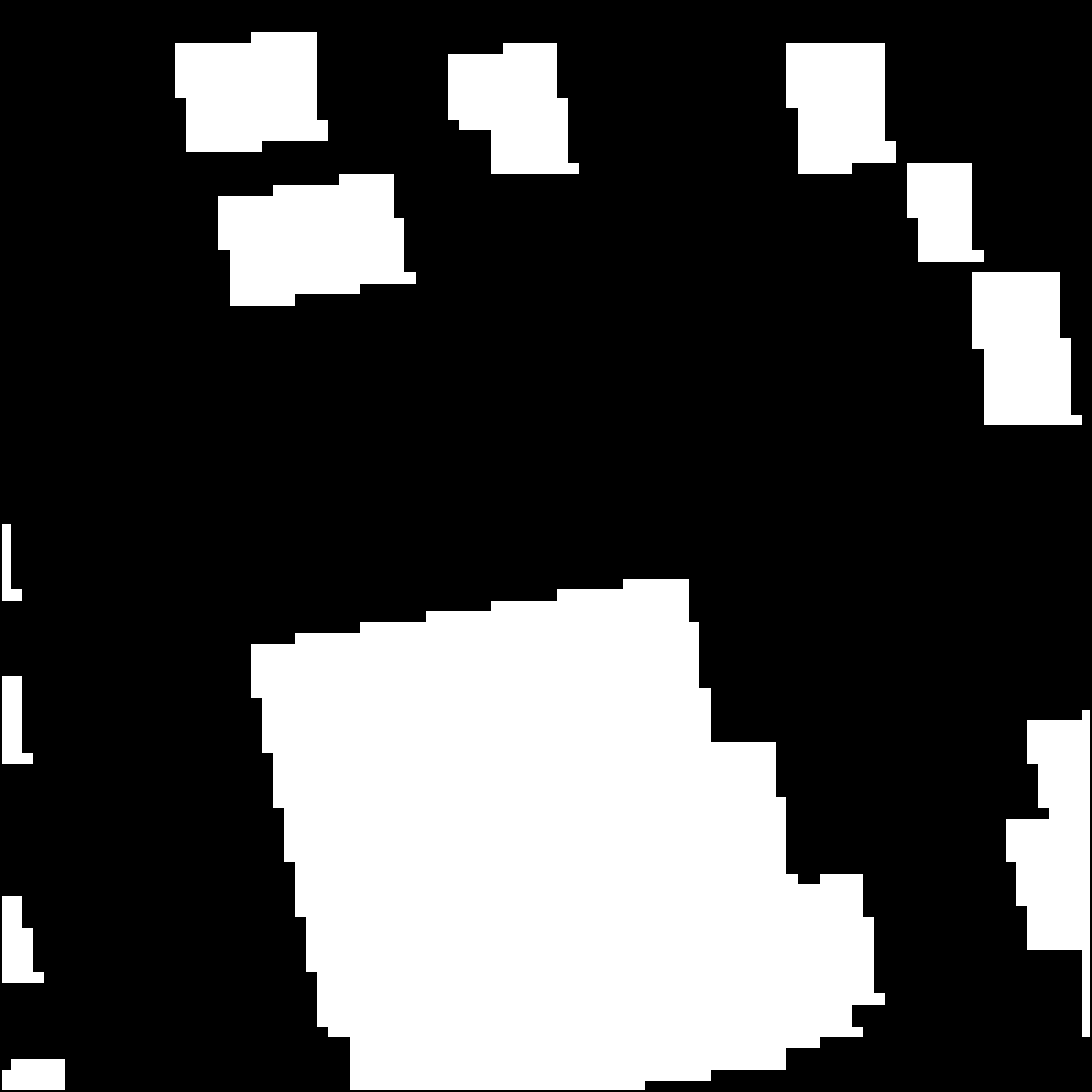}}
    \label{label_demo_a}
  \subfloat[Sign-distance transformed label]{%
        \includegraphics[width=4.2cm,height=4cm]{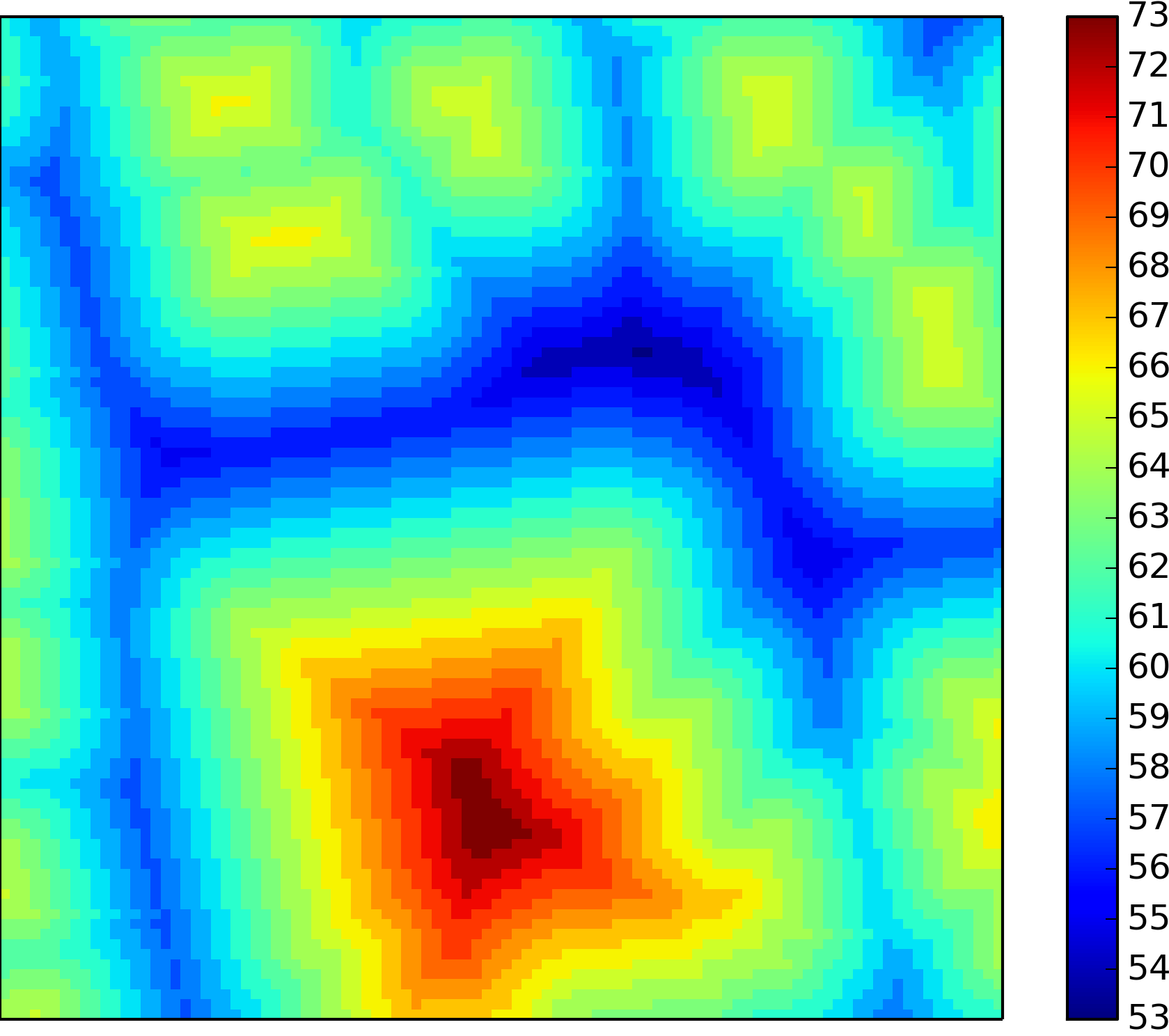}}
    \label{label_demo_b}
    \subfloat[Raw output of the trained network using signed distance labels]{%
        \includegraphics[width=4.2cm,height=4cm]{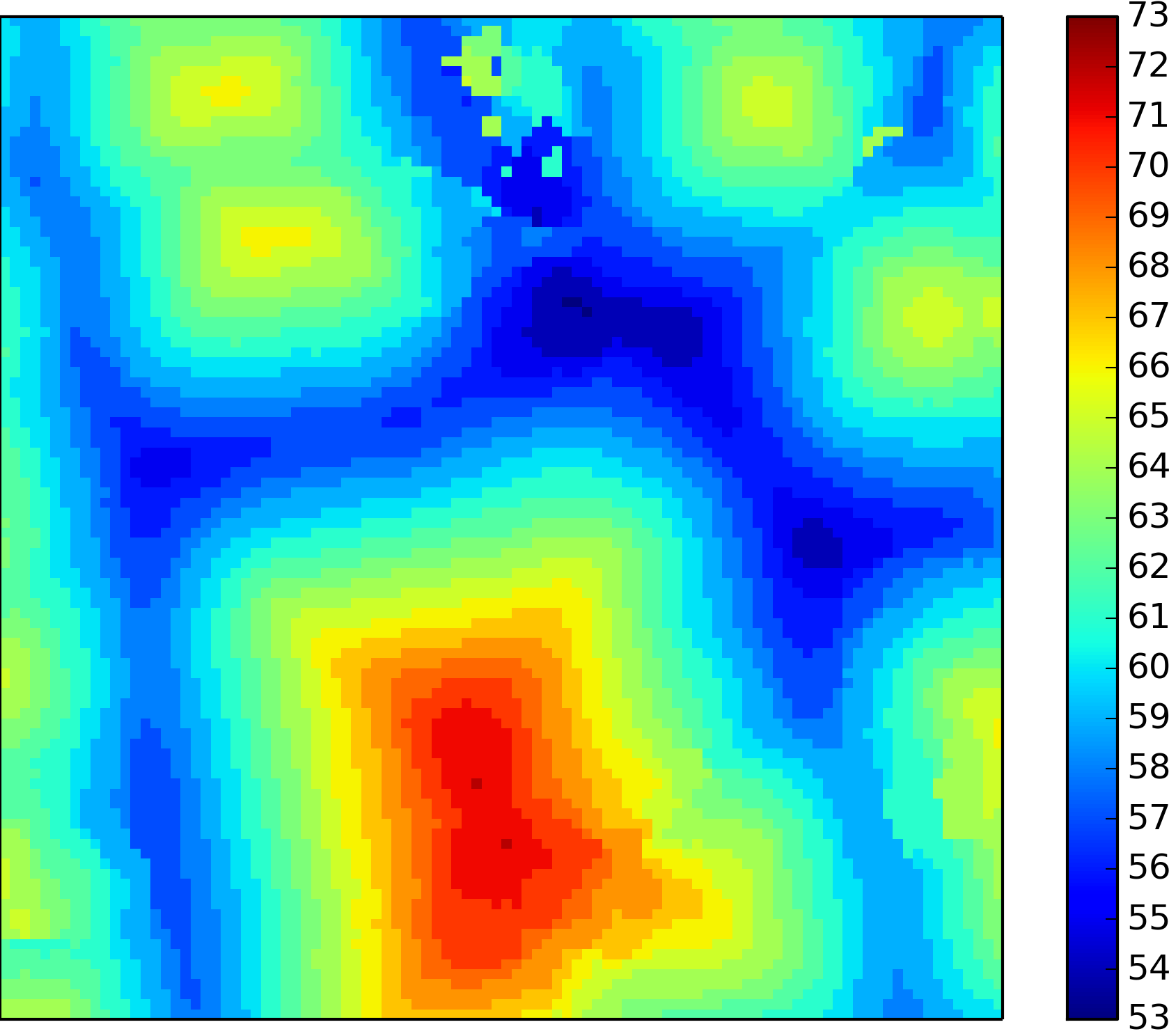}}
    \label{label_demo_c}
    \subfloat[Final output of extracted buildings based on (c)]{%
     \includegraphics[width=4cm,height=4cm]{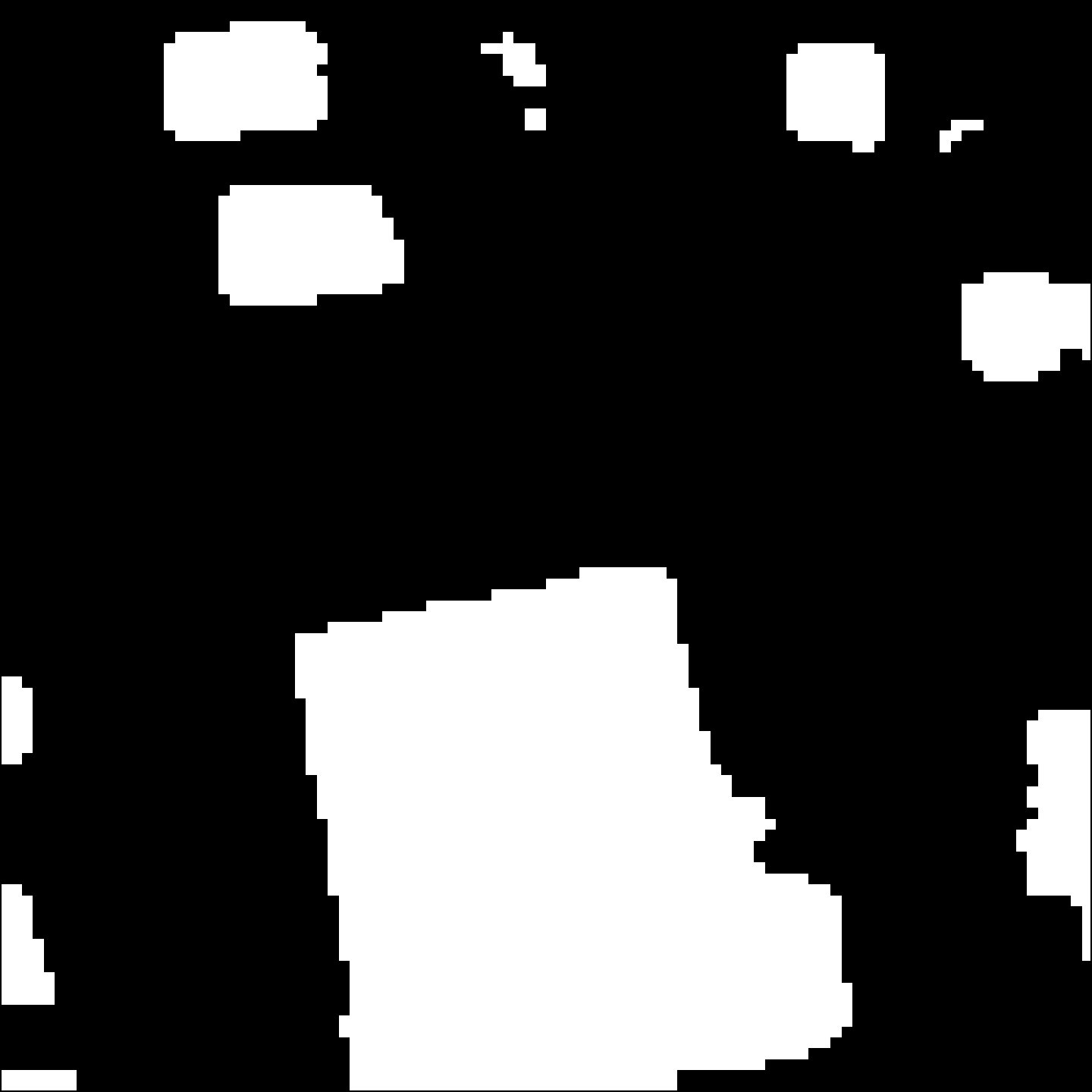}}
    \label{label_demo_d}
  \caption{Examples of binary label (a) and distance-transformed label (b), where building boundaries correspond to the class 64 in this example. The raw output of the trained network is shown in (c), where we can identify pixels with class label larger than 64 as buildings, as shown in the final building extraction result (d).  }
  \label{fig:label_demo} 
\end{figure}

Instead of using binary labels for building extraction, we propose to use the signed-distance function \cite{Yuan2018} to encode fine-grained information that present both the buildings and surrounding semantics (non-buildings). A similar concept has been proposed in \cite{Mnih2013} as soft labels, however, only considers the building class. Here we further take those surrounding objects of buildings into account: the signed-distance function value for a pixel is computed based on the distance from the pixel to its closest point on boundaries. We use positive values indicating inside of the buildings and negative otherwise. We can then cast our CNN learning problem with signed-distance labels as a multi-class classification problem by categorizing the distance values into a certain number of classes. In this paper, we define 128 classes with the labels $y=\{1,2,\dots,128\}$ where the first 64 classes all indicate buildings and the other 64 classes are for non-buildings. Since the signed-distance value is based on the distance from a given pixel to building boundaries, we set that the class 64 corresponds to building boundary and the pixels with the class labels larger than 64 are inside of buildings. Similarly, the pixels with the class labels smaller than 64 are defined as non-buildings. In Fig. \ref{fig:label_demo}, we can see the examples of a building footprint as in binary label, signed-distance transformed label, and also the output of a trained network using the signed-distance transformed label. With the network output in Fig. \ref{fig:label_demo} (c), we simply threshold the class labels to get the final building extraction result Fig. \ref{fig:label_demo} (d) :
pixels are buildings if $y>=64$ otherwise they are non-buildings: $y<64$.    

The signed-distance labeling provides additional information for CNNs to learn to distinguish different buildings based on their between-distance and labels. In addition, with binary labels, especially for SegNet, some sharp contrasts such as shadows in forests or roads in fields would often lead to inaccurate building extraction results. By incorporating surrounding semantics of buildings, we expect to reduce such errors.   

Given the buildings are sometimes occluded by trees, we also propose to exploit near infrared (near IR) band to suppress the impact of vegetation on building extraction results  \cite{JinDavis2005}. This is mainly motivated by the fact that near IR spectral characteristics of buildings and vegetation are distinct. Since we need to change the architecture of CNNs to take an additional band, directly using the available pre-trained models which mostly were trained with natural images taken from consumer-grade RGB cameras is not possible. Instead, 
we propose to use a simple fusing strategy to incorporate the near IR band into our building extraction work while still leveraging the power of pre-trained CNN models. 

First, we train one CNN model with RGB three channels images, and we also have another CNN model trained with CIR (IR, G, B) images. Note that CIR images were also successfully explored in \cite{Sherrah2016} and \cite{VakalopoulouKarantzalosKomodakisEtAl2015} for building extraction. 

In our case, there are two inferences for each pixel from the two CNN models, using majority voting or taking the average of the predicted labels obtained from softmax layers is not sufficient. We propose to fuse the likelihood results provided by the softmax layers with equal weights for each pixel and give the inference based on the averaged softmax results. Two models can be trained simultaneously, and generate predictions via fusion at the inference time. 

By considering the surrounding semantics of buildings and exploiting near IR information, we expect more accurate building extraction results both in class-level and instance-level. 
\section{Experiments}

Using cross validation metrics, we conduct extensive experimental analysis with current SOA CNNs for the task of pixel-labeling for building extraction. We improve upon the SOA results by two means (1) by incorporating the novel sign-distance labeling technique and (2) by considering different imagery inputs via a fused CNN framework. Based on the results, we will extract a single optimal CNN model that we deploy for large scale building mapping over the entire United States.

\subsection{Data Preparation}
To identify the availability of imagery with a reasonable spatial resolution is critical for building mapping in the United States. Fortunately, the United States Department of Agriculture's (USDA) National Agriculture Imagery Program (NAIP) provides 1-meter resolution imagery with four bands (red, green, blue, and near infrared) to the public. With consistently less than 10\% cloud coverage, the NAIP images cover the entire contiguous United States. For these reasons, we used NAIP images as our testbed imagery for scalable building extraction in the United States.

Currently, most building extraction research exploits data sets that are available to the public, such as 2D semantic labeling data sets provided by ISPRS \cite{VolpiTuia2017}. However, the fact that the performance of CNNs is sensitive to the characteristics of training data is one of our basis to motivate and argue that open source training data is neither sufficient nor suitable to achieve the goal of establishing large scale building extraction. Ideally, the differences between the training data and the to-be-processed NAIP images should be minimal to avoid dataset shifts phenomenon that leads to domain adaptation challenges \cite{Quionero-CandelaSugiyamaSchwaighoferEtAl2009}. Therefore, we need to seek alternative sources of training data to enable our large scale building mapping with CNN in the United States.

The building training samples were generated based on a LiDAR building footprints database. We compiled 5,173 500 by 500 pixels using 1-meter resolution NAIP images from nine cities. The nine cities were selected geographically spread across the United States. We randomly picked 4,000 images as the training set, and the remaining 1,173 images as the validation set.

\begin{figure}
	\includegraphics[scale=0.03]{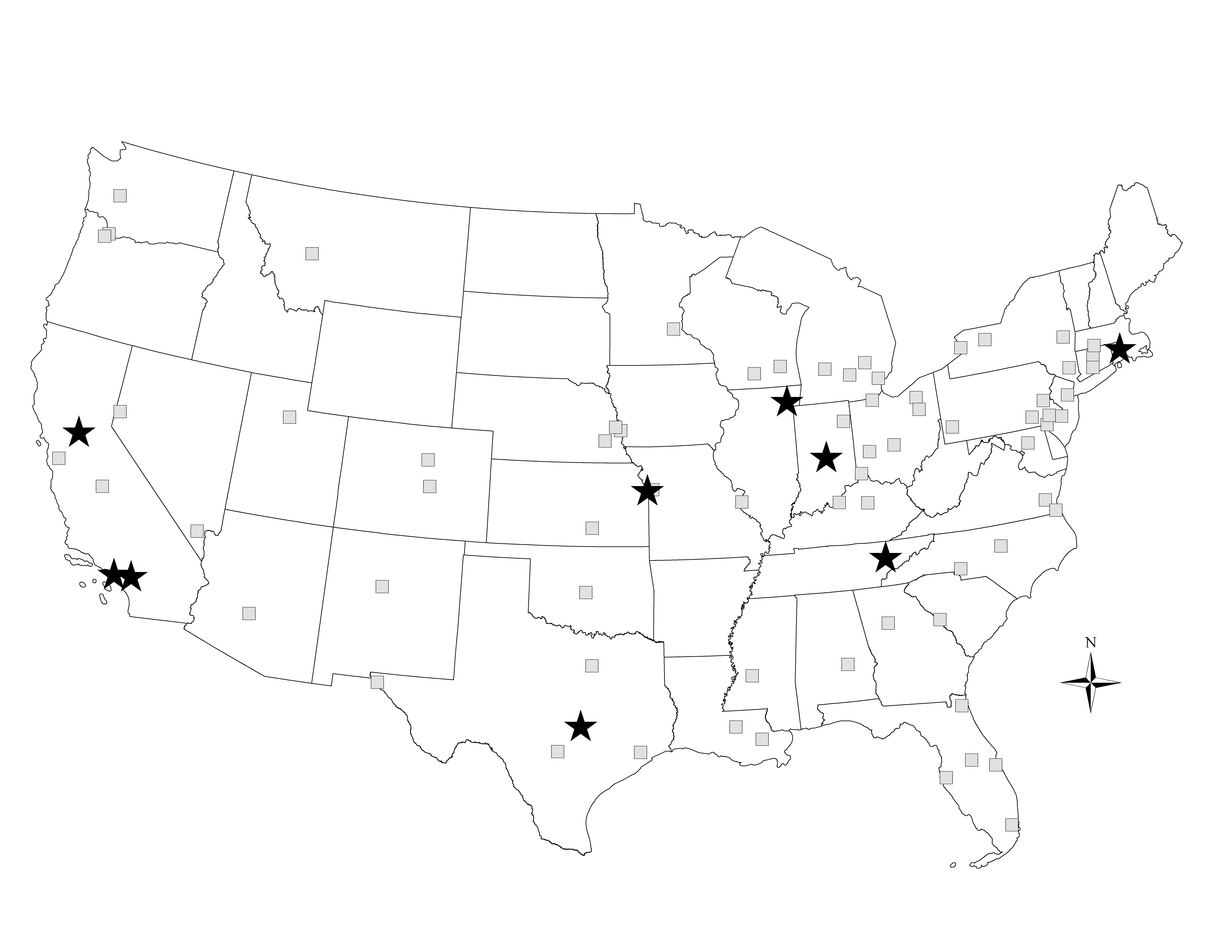}
\caption{Locations of training (stars) and testing sites (squares).}
\label{fig:loc_train_test}
\end{figure}

We picked a set of 78 image tiles covering sites different from training and validation data as the test set for performance evaluation. Each image tile is with size of $\sim$6000-by-$\sim$7000 pixels. Those testing sites are located across the entire continental United States. The locations of the training and testing sites are shown in Figure \ref{fig:loc_train_test}.  For each NAIP test tile there is a LiDAR building footprints map used as the ground truth. The LiDAR building footprint database covers certain major cities of the United States and is provided by National Geospatial-Intelligence Agency. It is worth noting that a NAIP image and its corresponding LiDAR building footprints were very likely collected at different dates, sometimes even in different years. Moreover, given the NAIP images are usually scheduled to collect data during the agricultural (leaf-on) season, the most common disagreement is observed when some of the buildings shown in the LiDAR database are covered by trees in the images. Therefore, we calculated the Normalized Difference Vegetation Index (NDVI) of the NAIP images and used it as the indicator to exclude those buildings shadowed by tall trees. In addition, we also need to consider the mis-alignment error between imagery and LiDAR building footprints due to different map projections and inevitable errors which occurred in the process of generating images and LiDAR building footprints. We follow the auto shifting procedure in \cite{YuanCheriyadat2014} to minimize the mis-alignment impact. Performing these two steps to prepare the ground truth data allows us to ensure the quality of training data, which has been indicated as an important factor to have a good results of CNN for building extraction \cite{MaggioriTarabalkaCharpiatEtAl2016}, and obtain reliable assessment. 

\subsection{Network Configurations}
The experimental settings for training each of the CNNs are briefly discussed in the following subsections.   
\subsubsection{FCN}
Using FCN, we conducted 8-stride (\textbf{FCN-8s}) as in \cite{ShelhamerLongDarrell2017} and  4-stride (\textbf{FCN-4s}) based experiments for the building extraction. In the FCN work \cite{ShelhamerLongDarrell2017}, 8-stride network is the finest resolution for the tested data sets. However, in our experiments, the 8-stride setting seemed too coarse as some of the buildings are not identifiable when one examines an image that is equivalently down-sampled 8 times. Based on this assumption, we also include a 4-stride FCN by fusing an extract detailed feature maps in the experiments. The networks were initialized with the pre-trained VGGNet \cite{Simonyan2015} model and were trained with batch size=1 as suggested in \cite{ShelhamerLongDarrell2017}. The stochastic gradient descent (SGD) algorithm was selected to solve the optimization problem in training CNN with the learning rate$=10^{-10}$, the weight decay parameter $=5\times10^{-4}$ and momentum$=0.99$. 

\subsubsection{CRFasRNN}
We connected the CRFasRNN to the trained FCN-8s and FCN-4s and obtained \textbf{FCN-8s-CRF} and \textbf{FCN-4s-CRF} models, respectively. We used the default CRF parameters as in \cite{ZhengJayasumanaRomera-ParedesEtAl2015}. We set the learning rate in SGD as $10^{-11}$ and momentum as $0.99$.
\subsubsection{SegNet}
Finally, in the SegNet experiments, with the same architecture we first initialized the pre-trained VGGNet \cite{Simonyan2015} model and used a batch size of 3 in training, which is the maximal size allowed based on our GPU capacity. We again used the stochastic gradient descent algorithm for training with the learning rate$=10^{-3}$, the weight decay parameter $=5\times10^{-4}$ and momentum$=0.9$. 
In the FCN and CRFasRNN experiments, we used  similar smaller learning rates to the oringial FCN and CRFasRNN papers since the loss used in SGD is summed spatially over all pixels, as noted in \cite{ShelhamerLongDarrell2017}.
We did not use the weighted loss strategy as in \cite{BadrinarayananKendallCipolla2017} to weight loss in the binary label experiments, as we found that the network tends to have higher false positive detection and requires longer training when weighted on the building class. For the experiments with signed-distance labels, we applied weighted loss function based on the frequency of each class similar in \cite{XieTu2015}.

All of our experiments to evaluate multiple CNN models  were carried out with a single NVIDIA-Tesla K80 GPU. The CNNs were implemented with CAFFE library and trained for 120,000 iterations. To generate a building map for the entire Continental United States the task is parallelized to perform inference across 8 NVIDIA-Tesla K80 GPUs.

\section{Results Discussion}
We evaluated the performance using four metrics. We used precision rate, recall rate, and F-score as our metrics for pixel-based evaluation. They are being used widely as the standard evaluations for building extractions \cite{Ngo2017,Ok2013}. We also included the results of the intersection over union (IoU) criterion, a popular metric for segmentation task \cite{ShelhamerLongDarrell2017,Maggiori2016}. Finally, we also included additional overall accuracy for completeness. The definition for these metrics is given in the following:

\begin{equation} \label{eq:precision}
\text{precision}= \frac{\text{TP}}{\text{TP}+\text{FP}}
\end{equation}
\begin{equation} \label{eq:recall}
\text{recall}= \frac{\text{TP}}{\text{TP}+\text{FN}}
\end{equation}
\begin{equation} \label{eq:f-score}
\text{F-score}= \frac{2\cdot\text{precision}\cdot\text{recall}}{\text{precision}+\text{recall}}
\end{equation}
\begin{equation} \label{eq:IOU}
\text{IoU}= \frac{\text{TP}}{\text{TP}+\text{FP}+\text{FN}}
\end{equation}
\begin{equation} \label{eq:OA}
\text{Accuracy}= \frac{\text{TP}+\text{TN}}{\text{TP}+\text{FP}+\text{FN}+\text{TN}}
\end{equation}
where \textbf{TP} denotes true positives (correctly extracted building pixels), \textbf{FP} denotes false positives (pixels mislabeled as buildings in results), \textbf{TN} denotes true negatives (correctly identified non-building pixels), and \textbf{FN} denotes false negatives (pixels incorrectly labeled as non-buildings or can be interpreted as missed building pixels). \\
\subsection{Training CNN Models with Binary Labels}
\begin{figure*}[t!]
	\includegraphics[scale=0.34]{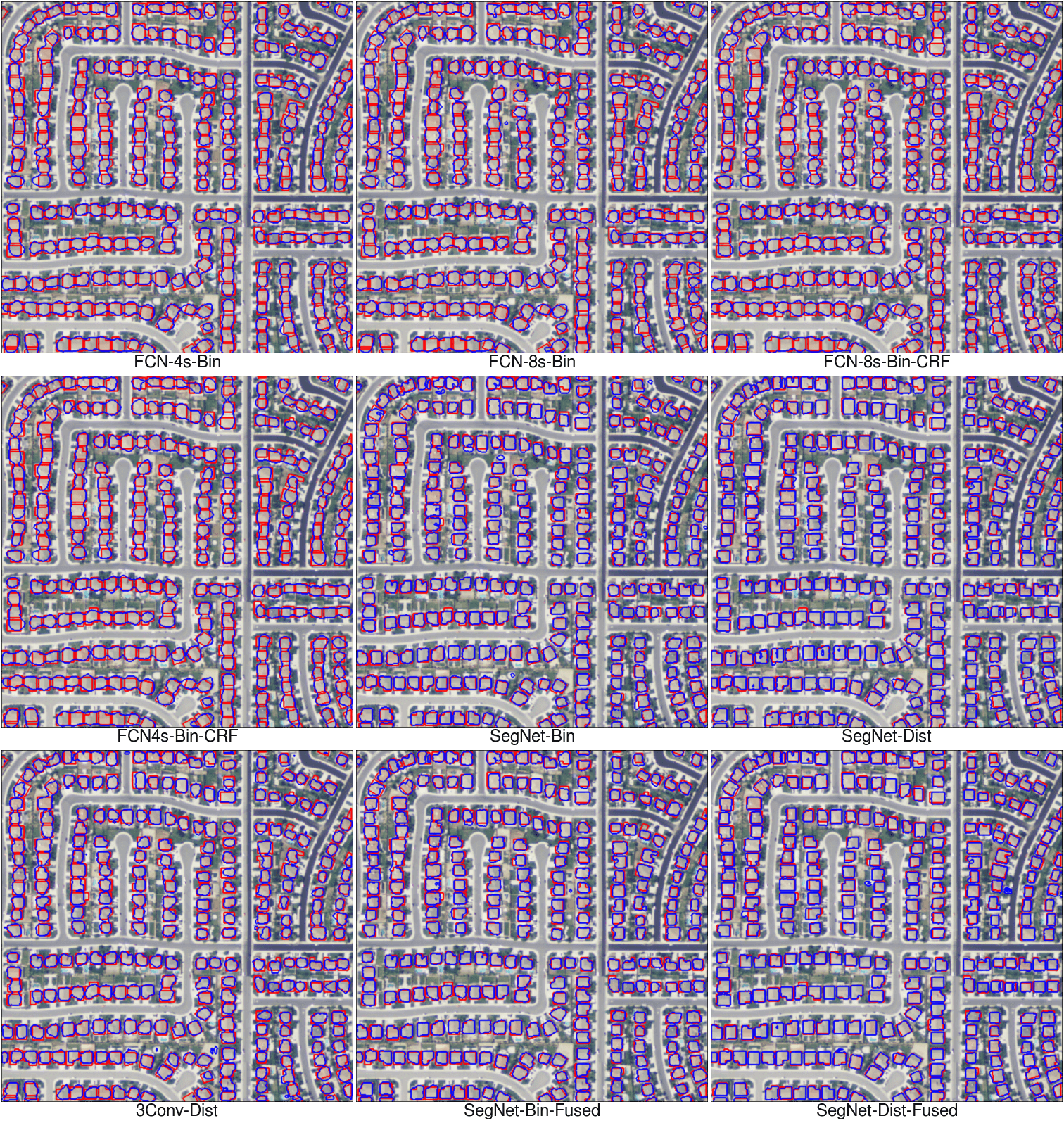}
	\caption{Example: High density building area where red lines delineate the  building extraction results and blue lines denote the ground truth.}
	\label{fig:res_high_den_building}
\end{figure*}

We first tested different CNNs with binary labels in this section to evaluate their applicability for building extraction. The averaged precision, recall, F-score,IoU derived from the 78 sites results using binary labels are shown in the Table \ref{tab:naip_78sites_acc} with the suffix \textbf{Bin}. First, we noted that \textbf{FCN-4s-Bin} and \textbf{FCN-8s-Bin} provided comparable results in these metrics. The observation implies an extra input from higher resolution feature maps contribute minimal to improve the results. As pointed out in \cite{ZhengJayasumanaRomera-ParedesEtAl2015, BadrinarayananKendallCipolla2017}, the boundaries of objects provided by FCN are imperfect and sometimes lacking. We also observed the similar issue in our building extraction results of FCNs, especially in urban areas where there are clusters of smaller buildings with minimal separations. The extraction results provided by FCNs tend to be blob-like structures, as shown in Figure \ref{fig:res_high_den_building}. In this example, the community-style buildings are essentially identified as a large object for those nearly connected buildings. In this case, using FCN as a special treatment cannot effectively compensate the reduced spatial details resulted from convolution and pooling operations, even with FCN-4s. 

\begin{figure}[htb]
	
	\begin{minipage}[b]{0.46\linewidth}
		\centering
		\centerline{\includegraphics[width=1\linewidth]{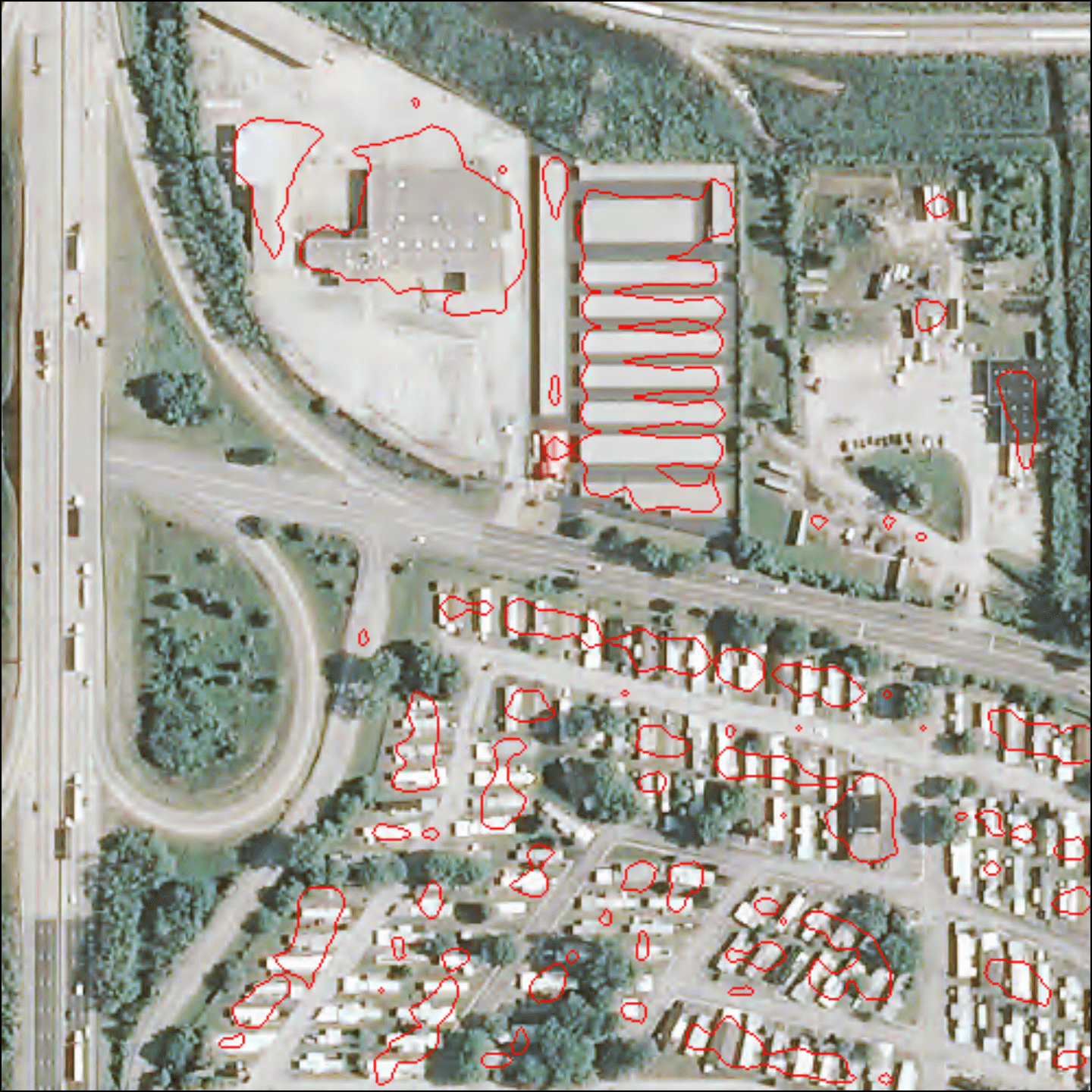}}
		\centerline{(a) FCN-8s-Bin}\medskip
	\end{minipage}
	\begin{minipage}[b]{0.46\linewidth}
		\centering
		\centerline{\includegraphics[width=1\linewidth]{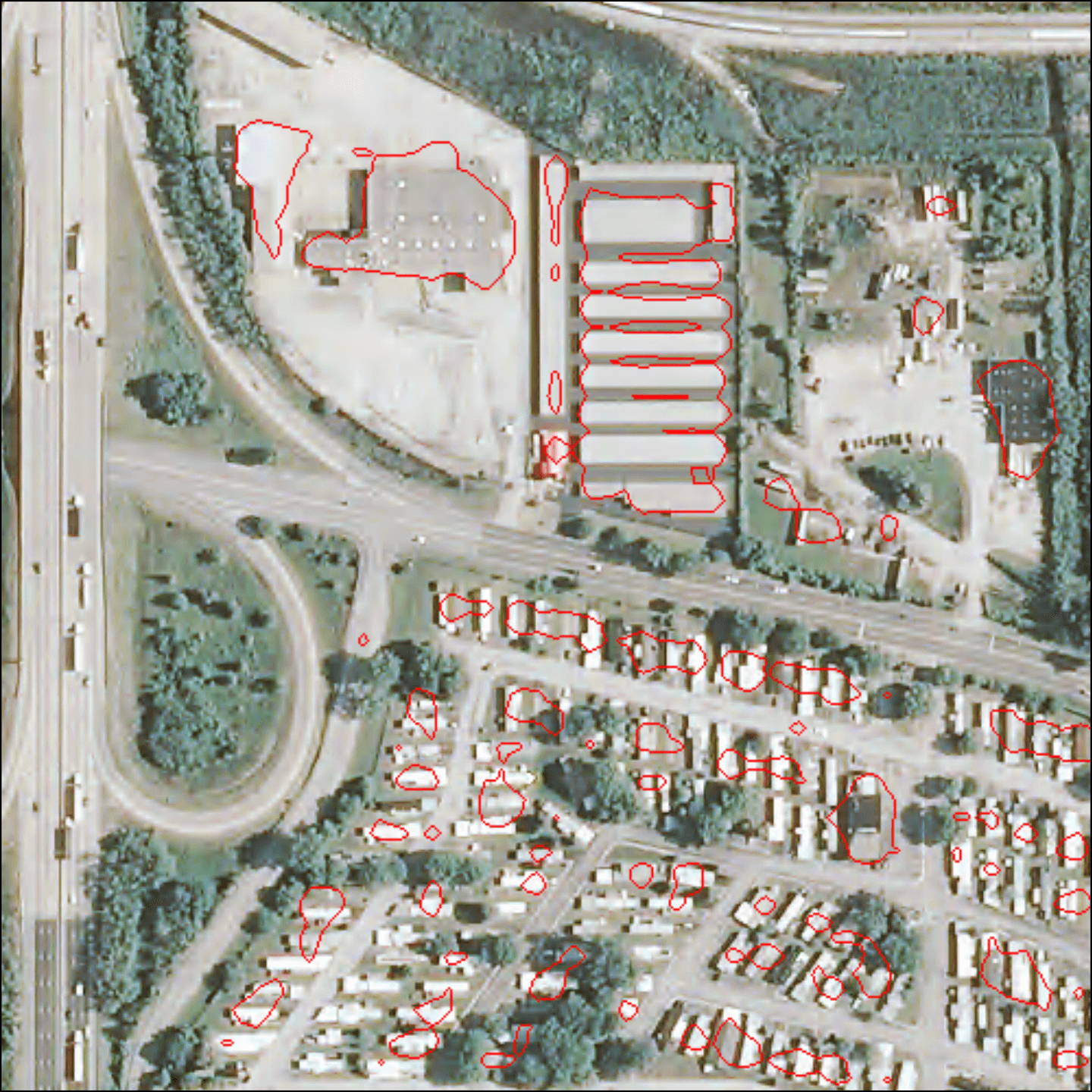}}
		\centerline{(b) FCN-8s-CRF-Bin}\medskip
	\end{minipage}

	\caption{\label{fig:fail_fcn}An example of indistinguishable building outlines for FCN-8s-Bin and FCN-8s-CRF-Bin. Red lines delineate the  building extraction results.}

\end{figure}

We then evaluate the benefits of using the conditional random field plug-in to refine the building boundaries. As listed in Table \ref{tab:naip_78sites_acc}, only the \textbf{FCN-8s-CRF-Bin} results show more improvements in terms of recall and F-score. Our further investigation indicates the performance of CRF module is sensitive to the initial extractions provided by FCNs. Figure \ref{fig:fail_fcn} presents a visual example on a challenging building extraction scene that concrete parking lots and pavements are close to buildings. This scenario presents a more formidable task to correctly identify building outlines when the image contrast is inferior. Compared to Figure \ref{fig:fail_fcn} (a), Figure \ref{fig:fail_fcn} (b) demonstrates the advantages of incorporating CRF to improve the detected building boundaries in this case. For example, the outlines of the black building on the right and the gray building in the middle have been refined. However, if the spacing between buildings is small, as discussed earlier, the result provided by the FCN family generally is a large connected object and using CRFasRNN yields minimal improvements. It is possible to improve further with more iterations with tuned learning parameters of CRF. However, it implies longer training time in a deep CNN and efforts to find optimal CRF parameters.

\begin{figure}[t!]
	\includegraphics[scale=0.12]{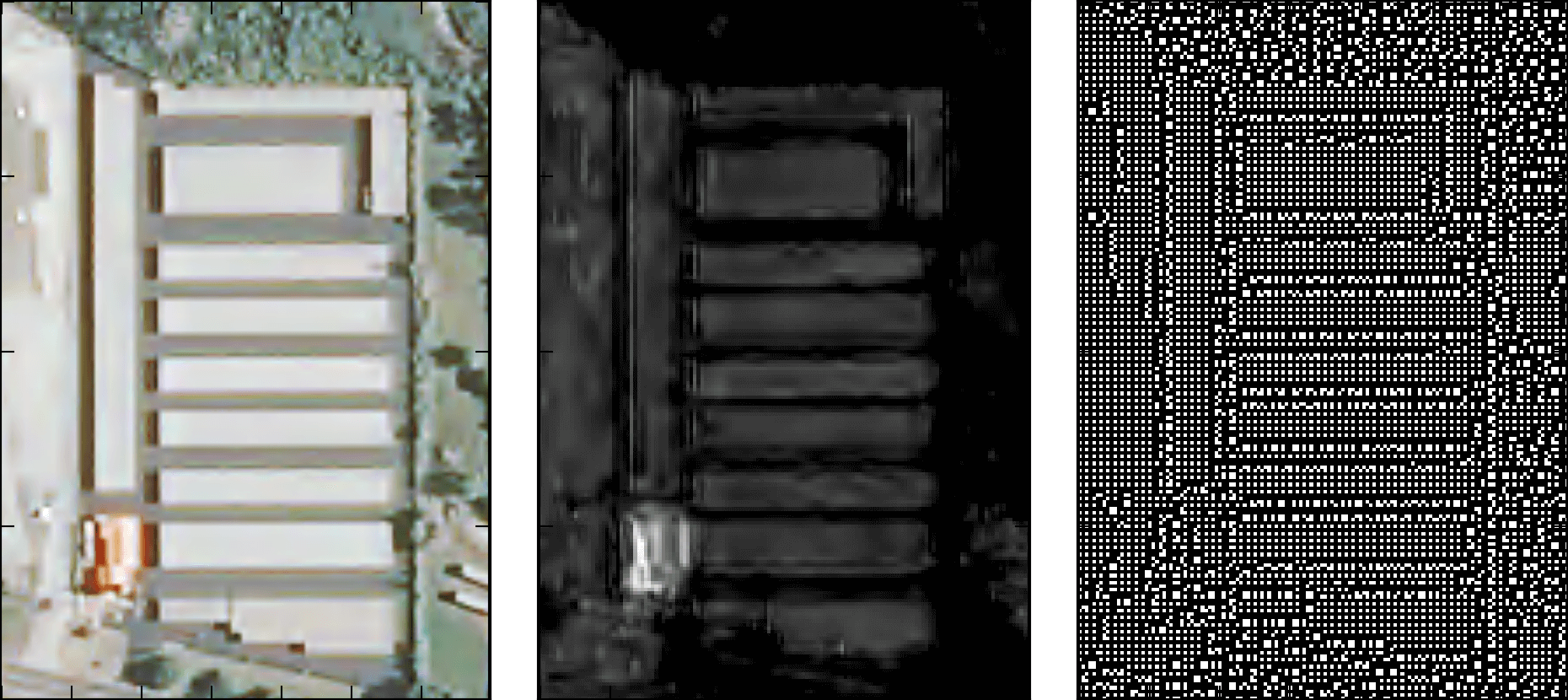}
	\caption{The input RGB image (left), one of the feature maps (central) and the corresponding max-pooling indices (right). We can also observe the right figure features a map with sparse white dots, whose locations correspond to the max-pooled indices.} 
	\label{fig:segnet_bin_index_ex}
\end{figure}

The advantages of using the indices of max-pooling in SegNet to capture the strong features corresponding to edges of building is demonstrated quantitatively in Table \ref{tab:naip_78sites_acc} where \textbf{SegNet-Bin} provides the best F-score (0.68) and IoU (0.52) among all models trained with binary labels. As the sample shown in Figure \ref{fig:segnet_bin_index_ex}, the sparse indices map to those essential features of buildings (edges, corners), the central one is one of the convoluted feature map provided by the first set of convolution operations, and the right one shows the locations (white dots) that corresponds to the max-pooled pixels. We also note that the individual buildings can be extracted with \textbf{Seg-Bin} in Figure \ref{fig:res_high_den_building}. 

\begin{table*}[htbp]	
	\centering
	
	\caption{Building extraction results of the disjoint 78 testing sites}
	
	\begin{tabular}{lccccc}
		
		\hline\hline
		
		Method & \multicolumn{1}{l}{Precision} & \multicolumn{1}{l}{Recall} & \multicolumn{1}{l}{Accuracy} &\multicolumn{1}{l}{F-score} & \multicolumn{1}{l}{IoU} \\
		
		\hline\hline
		
		FCN-4s-Bin & 0.75  & 0.54  & 0.95&0.62  & 0.46 \\
		
		FCN-8s-Bin & 0.75  & 0.55  & 0.95& 0.62  & 0.46 \\
		
		\hline
		
		FCN-4s-CRF-Bin & 0.74  & 0.55  & 0.95&0.62  & 0.45 \\
		
		FCN-8s-CRF-Bin & 0.74  & 0.58  & 0.95&0.64  & 0.48 \\
	
		\hline
		
		SegNet-Bin & 0.79  & 0.61  & 0.96 & 0.68  & 0.52 \\
		
		SegNet-Dist & 0.77  & 0.66  & 0.96& 0.71  & 0.55 \\
		\hline
		3Conv-Dist&0.73&0.67& 0.95&0.69&0.53\\
		\hline
		SegNet-Bin-Fused &0.81 &0.63&0.96&0.70&0.55\\
		SegNet-Dist-Fused& 0.73 & 0.74& 0.96&0.73& 0.58\\
		\hline\hline
		
	\end{tabular}%
	
	\label{tab:naip_78sites_acc}%
	
\end{table*}%

\begin{figure*}
	\includegraphics[scale=0.34]{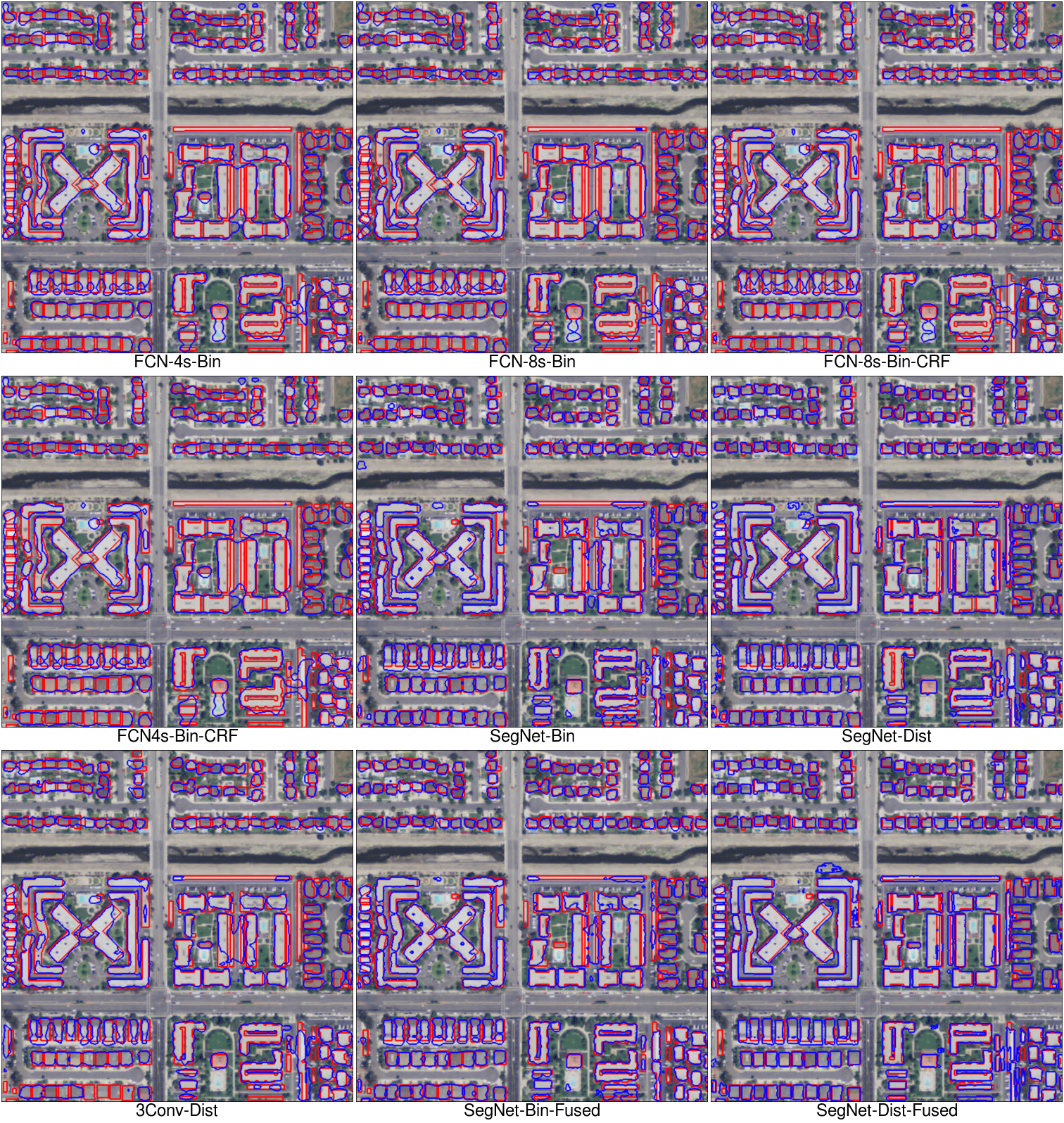}
	\caption{Example: Complex building area where red lines delineate the building extraction results and blue lines denote the ground truth.}
	\label{fig:res_complex}
\end{figure*}

\subsection{With Signed-distance Labels and Model Fusion}
After reviewing the results provided by the CNN models trained with binary labels in Section V. A., we then investigate the effectiveness of incorporating signed-distance function for building extraction. Since the SegNet CNN with binary labels \textbf{SegNet-Bin} significantly outperforms other CNNs, we selected the SegNet model as the base CNN architecture to study the usefulness of signed-distance labels. The relavant results are listed in Table \ref{tab:naip_78sites_acc} with suffix Dist. In addition, we included the results of the CNN model in \cite{Yuan2018} \textbf{3Conv-Dist} where the signed-distance function is used in a branch out style CNN with 3 convolutional layers and 3 upsampling layers. The implementation details of \textbf{3Conv-Dist} can be found in \cite{Yuan2018}.

Although the results of \textbf{SegNet-Bin} are superior to other methods with binary labels in terms of the capability of distinguishing individual buildings and better results in the four metrics, with the signed-distance labels, the trained model \textbf{SegNet-Dist} presents even more improvements from \textbf{SegNet-Bin} as shown in Table \ref{tab:naip_78sites_acc}. As an illustration, the \textbf{SegNet-Dist} model can differentiate individual small buildings accurately on the top part of the example image in Figure \ref{fig:res_complex}.

To further validate the correlation between signed-distance labels and effectiveness of extracting buildings at instance level, we show the percentage of the individual buildings detected in Table \ref{tab:detRate}. The detection of a single building is characterized by a single connected polygon (i.e. an extracted object) which (may) cover several small buildings i.e. counting the number of building we only account for the connected polygon as one building. In this way, we can only evaluate fairly if the buildings are extracted at instance level. With this case, the pixel-based precision and recall rate will still be high, as those pixels inside in the large object are indeed classified as buildings correctly.
We first note that with much fewer layers, \textbf{3Conv-Dist} provides a significantly higher percentage of buildings detected than those complicated, deeper CNNs trained with binary labels (\textbf{FCN-4s-Bin}, \textbf{FCN-8s-Bin}), which indicates the potential benefits of using signed-distance labels for building extraction. However, fewer buildings are detected with \textbf{3Conv-Dist} compared to \textbf{SegNet-Dist}, which confirms the effectiveness of our proposed strategy: combining max-pooling indices and signed-distance function for accurate building extraction.

The fused results of \textbf{SegNet-Bin} and \textbf{SegNet-Dist} are also listed in \ref{tab:naip_78sites_acc} and \ref{tab:detRate}, denoted as \textbf{SegNet-Bin-Fused} and \textbf{SegNet-Dist-Fused}, respectively. Both of the fused models yield better results than the ones obtained only with RGB bands. \textbf{SegNet-Dist-Fused} performs the best in terms of precision, recall, F-score, IoU and the number of buildings detected among the tested nine models, as shown in Table \ref{tab:naip_78sites_acc} and \ref{tab:detRate}. We further examine the results provided by CIR input. With binary labels and CIR input, the trained model yields precision 0.75 and recall 0.65 whereas using signed-distance labels with CIR data provides precision 0.74 and recall 0.68. Compared to the results obtained with RGB inputs (shown in Table  \ref{tab:naip_78sites_acc}), we see two sets of inputs show the distinctive advantages on improving precision and recall performance, respectively. With the simple linear fusing strategy, the promising performance demonstrated by fusing two models indicates that exploring near IR spectral band is indeed beneficial to the building extraction task. Although this conclusion is not surprising, we believe that further work focusing on co-training with all bands in one model will provide better results.  
\begin{table}[ht!]
	\small
	\begin{tabular}{ccc}
		\hline\hline
		FCN-4s-Bin & FCN-8s-Bin & FCN-4s-CRF-Bin \\ \hline
		31.0\% & 34.5\% & 29.8\% \\ \hline\hline
		 FCN-8s-CRF-Bin & SegNet-Bin &SegNet-Dis \\ \hline
		 35.4\% &74.1\% &83.7\% \\ \hline\hline
		  3Conv-Dist & SegNet-Bin-Fused & SegNet-Dist-Fused\\ \hline
 57.6\%&74.2\% &84.9\%\\

		\hline
	\end{tabular}		
	\caption{The percentage of the buildings detected by the nine tested models. }\label{tab:detRate}
\end{table}

\begin{figure}[tp!]
	\includegraphics[scale=0.3]{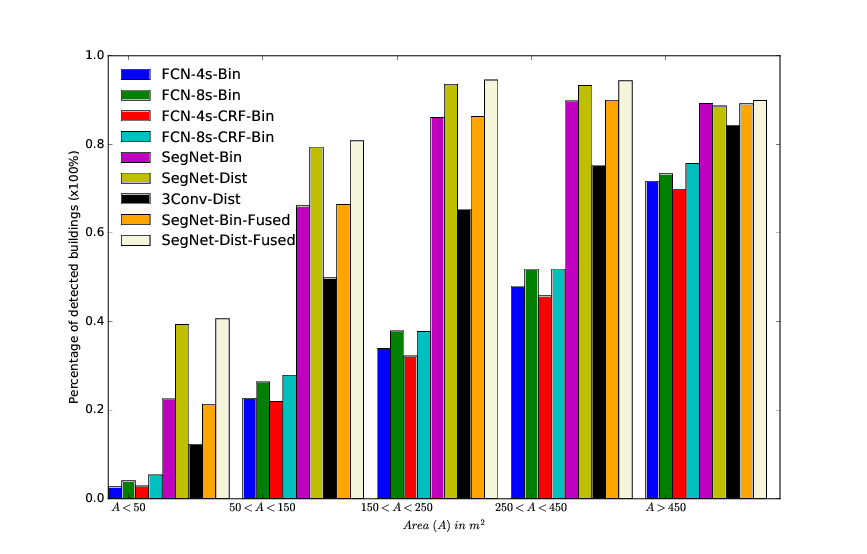}
	\caption{Various size of building detected by nine models.}
	\label{fig:no_building_detected_area}
\end{figure}
\begin{figure*}[t!]
	\includegraphics[scale=0.34]{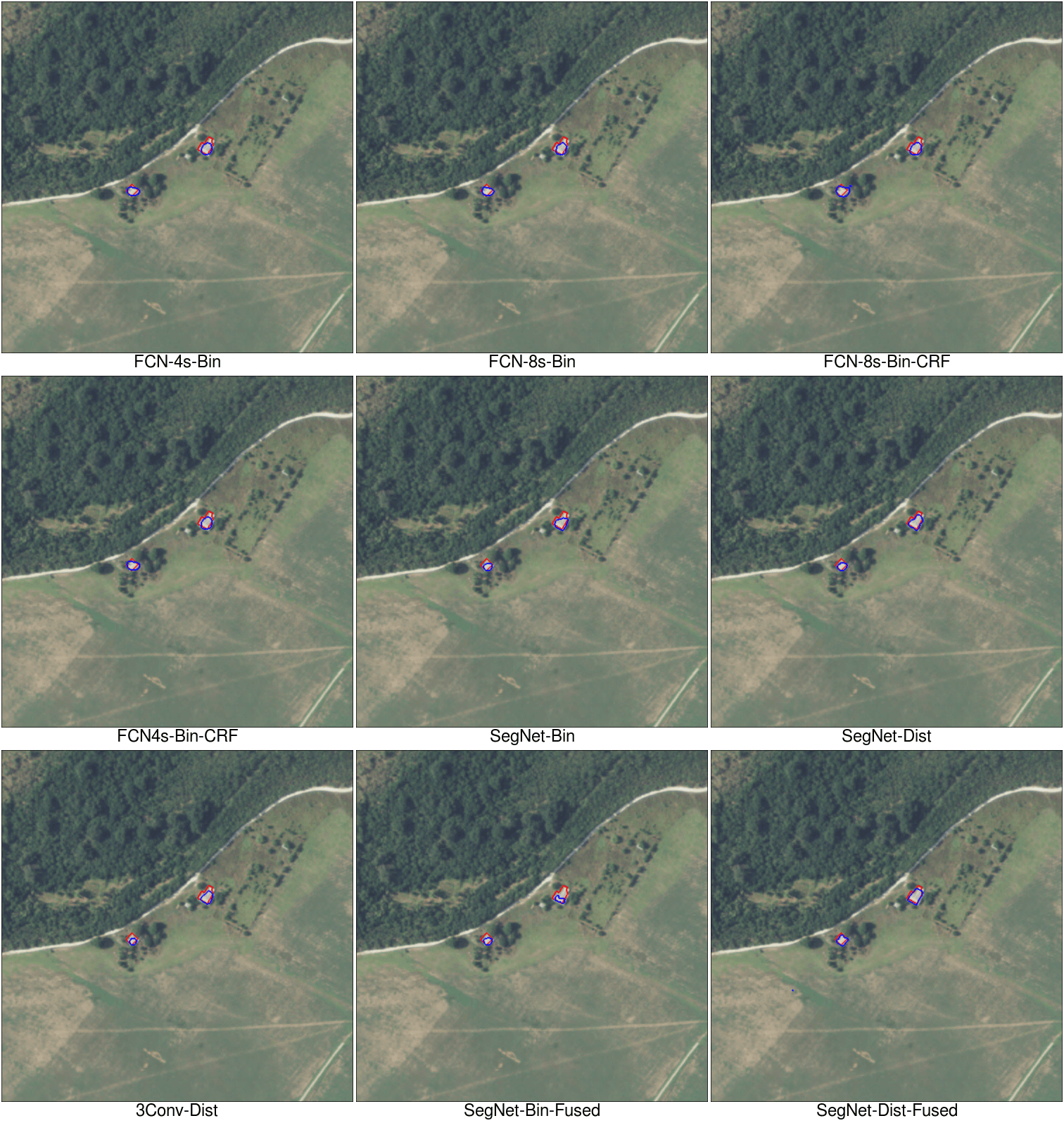}
	\caption{Example: Agriculture area where red lines delineate the  building extraction results and blue lines denote the ground truth.}
	\label{fig:res_ag}
\end{figure*}
The significant increase observed in the percentage of building detected with SegNet based methods can further be examined by relating the size of buildings to the number of buildings detected. We divided all buildings in the ground truth database into five groups based on their area: 1) Area (A) $\leq 50 m^2$, 2) $50m^2<$ A$\leq 150m^2$, 3) $150m^2<$A$\leq 250m^2$, 4) $250m^2<$A$\leq 450m^2$ and 5) $A>450m^2$. As shown in Figure \ref{fig:no_building_detected_area}, there are clearly two clusters for each group: The FCN and FCN-CRF based methods generally provide less impressive number of detected buildings, especially for smaller buildings (in group 1, 2 and group 3); the SegNet based methods, by contrast, show significant advantages in those groups. The boosted performance in all groups provided by signed-distance labels combining with max-pooled indices again confirms our proposed strategy to advance building extraction results to instance level.

We also provide the three example results provided by the nine methods: a high density building area in Figure \ref{fig:res_high_den_building}, some examples of complex building structures in Figure \ref{fig:res_complex}, and a agriculture area with sparse buildings in Figure \ref{fig:res_ag}. Due to limited space, we are constrained to show more examples over extended area. However, the advantages and disadvantages of each method we discussed above generally can be visually examined in these three examples that cover three common scenarios in our testbed dataset.

\begin{figure}
	\includegraphics[width=1.0\textwidth]{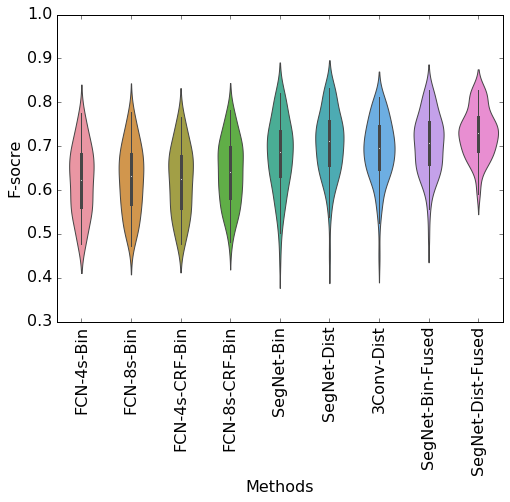}
	\caption{The violin plot of the F-score}
	\label{fig:violinplot_fscore}
\end{figure}

\begin{figure}
	\includegraphics[width=1.0\textwidth]{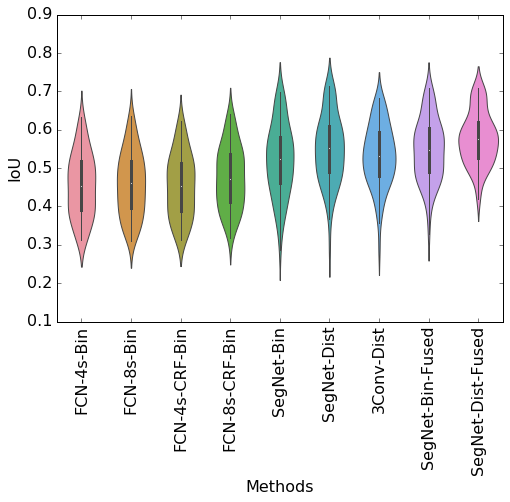}
	\caption{The violin plot of the IoU}
	\label{fig:violinplot_iou}
\end{figure}

\begin{figure}
	\includegraphics[width=1.0\textwidth]{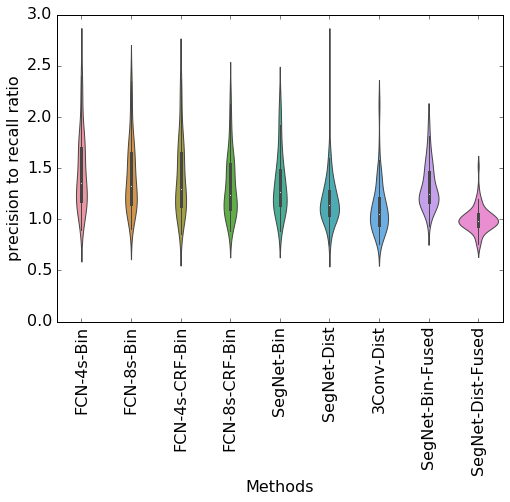}
	\caption{The violin plot of the ratio of precision and recall rates}
	\label{fig:violinplot_p2rRatio}
\end{figure}

In addition to the averaged metrics, we also would like to understand the variations in the performance of the 78 testing sites. As our goal is to apply a trained model to the contiguous United Sates, the desired outcome of a given model should be as consistent as possible across the 78 sites. If large variations of the corresponding results are observed for a given model, the model might be not suitable for large scale building extraction, since the inferior generalization is implied. To investigate the consistency of the results over 78 sites, we use violin plots in which the box plot and kernel density estimation are combined to illustrate the metrics obtained from each site. In Figure \ref{fig:violinplot_fscore}, we can see that F-score for the \textbf{Seg-Dist-Fused} yields the most uniform results across the 78 sites. Similarly, we can also conclude that the \textbf{Seg-Dist-Fused} gives the most consistent IoU for all testing sites as shown in Figure \ref{fig:violinplot_iou}. We also defined a simple precision-to-recall ratio and plotted a corresponding violin plot in Figure \ref{fig:violinplot_p2rRatio}. This ratio provides insights on the trade-off between precision and recall rate for each site as we desire the commission error (associated with precision rates) and omission error (associated with recall rates) would be balanced. That is, an ideal model should not be skewed toward buildings-sensitive or nonbuildings-sensitive. As shown in Figure \ref{fig:violinplot_p2rRatio}, the fused strategy clearly has the advantage to provide equally better precision and recall accuracy for all testing sites, noting the peaks around 1 for both fused strategies. Also, we note that most of the models are prone to false positive errors (larger ratios). Applying the signed-distance function to generate labels for training \textbf{SegNet} helps correct the skewed model, as seen in the increased peaks in \textbf{SegNet-Dist} v.s. \textbf{SegNet-Bin} and \textbf{3Conv-Dist}. Fusing another model trained with the additional IR band further balances the true positive and false positive detection.


\section{Generating Country Scale Building Maps and Identifying Challenges}
\begin{figure*}
    \centering
  \subfloat[]{%
       \includegraphics[scale=0.08]{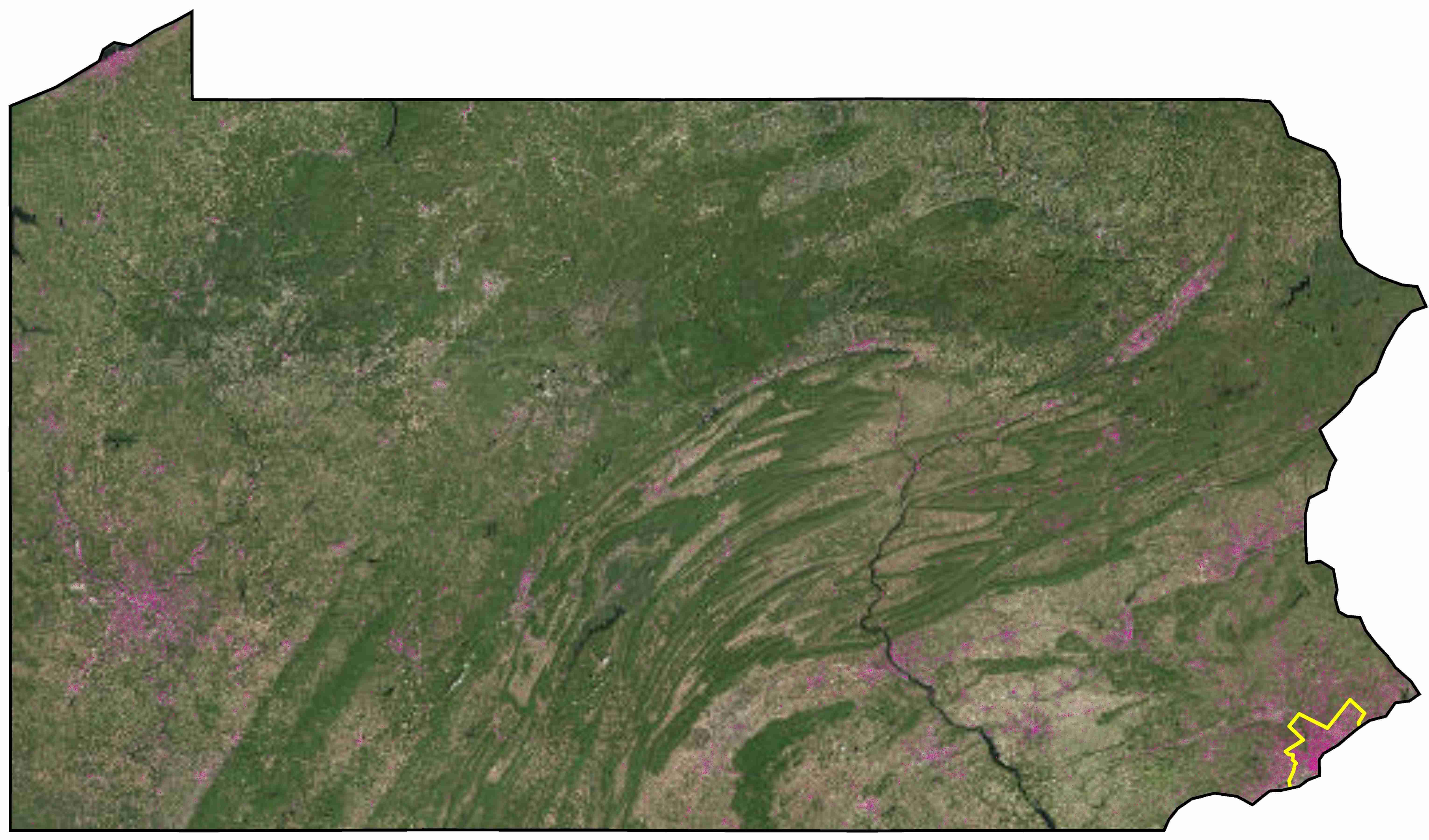}}
    \label{pa_state}
  \subfloat[]{%
        \includegraphics[scale=0.2]{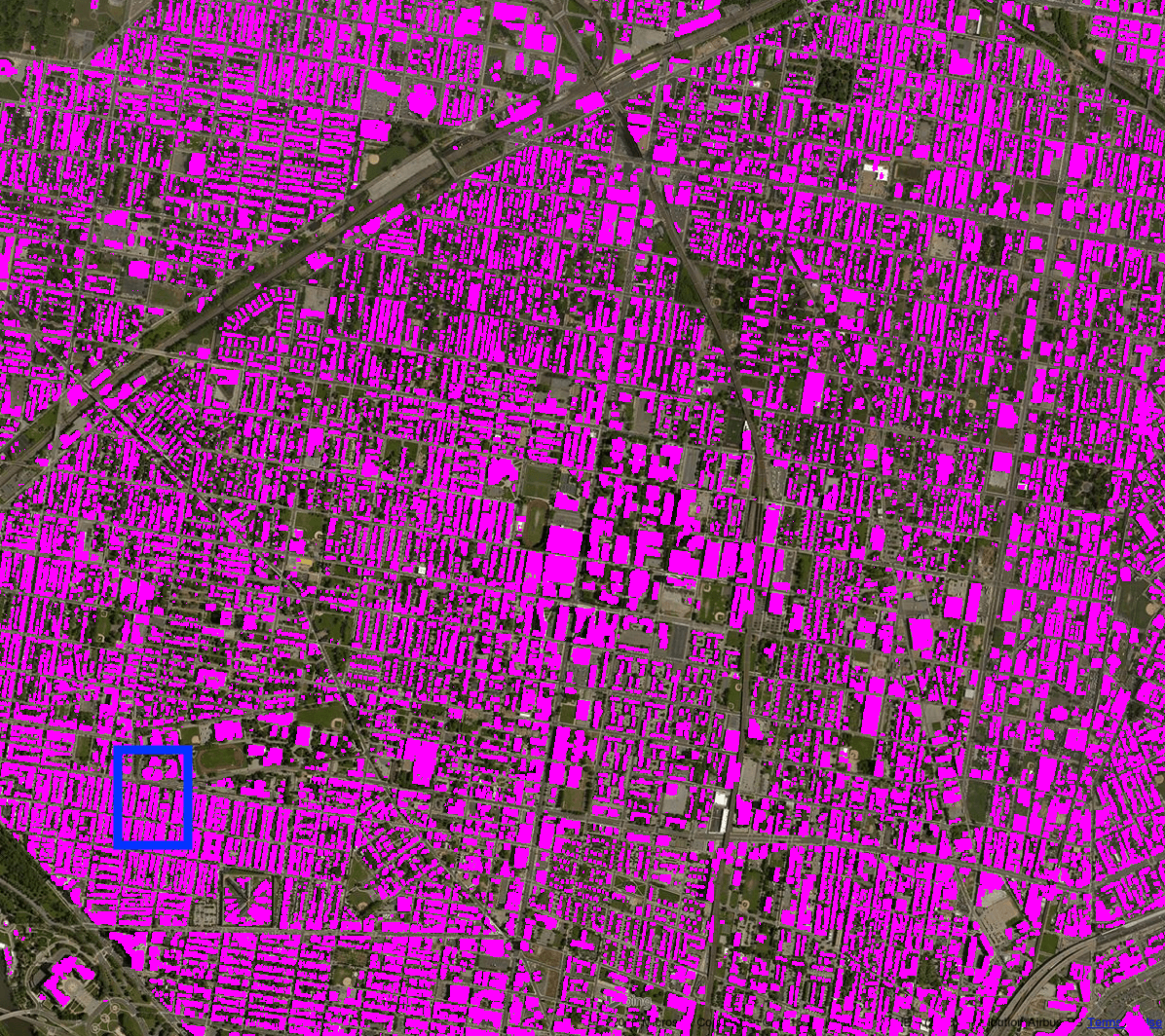}}
        \label{pa_county}
  \subfloat[]{%
        \includegraphics[scale=0.2]{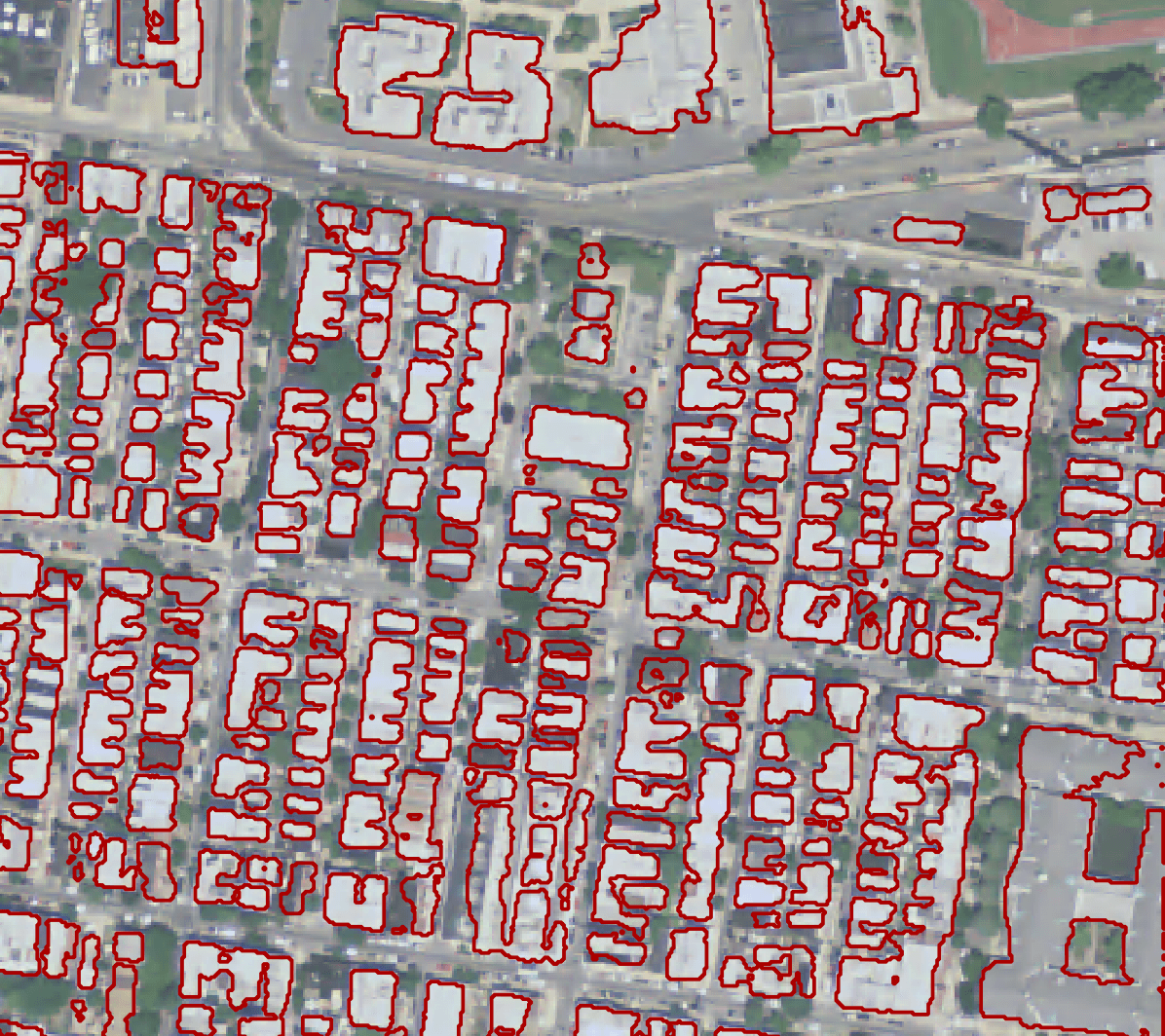}}
        \label{pa_city}
  \caption{The building extraction results in magenta for (a) the state of Pennsylvania, (b) the city area of Philadelphia county, which is the yellow polygon in (a). The red lines in (c) delineate the extracted building outlines over the downtown area of Philadelphia (blue box in (b))}
  \label{fig:pa_result} 
\end{figure*}

With the preferred model, we processed all NAIP images and established the first building maps covering the contiguous United States with an average processing time less than one minute for an area of size $\sim56$ $km^2$ per each of 8 NVIDIA Tesla K80 GPUs. All of the building extraction results are raw outputs from the model without post-processing. Due to space limit, we can only provide a visual example of the large scale building extraction result at state level. The results for the state of Pennsylvania are shown in Fig. \ref{fig:pa_result} (a),  and more zoom-in results at county level and city level are also provided in \ref{fig:pa_result} (b) and \ref{fig:pa_result} (c), respectively. Note that we do not have training data for this state and the trained model can still provide satisfying building extraction results.

Upon performing quality check state by state, although the building extraction results agree to the actual buildings seen in the images for most of the states, we found that the largest challenge to obtain the same performance level for every state is associated with the image quality and radiometric characteristics of the imagery. The image quality issue stems from the fact that NAIP images for each state are collected by multiple contractors and are not likely with the same camera, which inherently generates variations within the NAIP imagery. In some states, we noted that the corresponding images tend to be more blurry than other states and with lower color contrast. As a result, the performance of building extraction for these states are relatively poor. In addition, some of the states are largely dominated by desert or mountainous terrain and present quite unique landscape as compared to those included in training data. Although the pixel values of the four bands in NAIP images do not translate to spectral characteristics directly without proper radiometric calibration, we still can gain insights into such distinction by looking into the varied radiometric characteristics resulted from different landscapes. 

\begin{figure}[htb]
	
	\begin{minipage}[b]{0.49\linewidth}
		\centering
		\centerline{\includegraphics[width=1\linewidth]{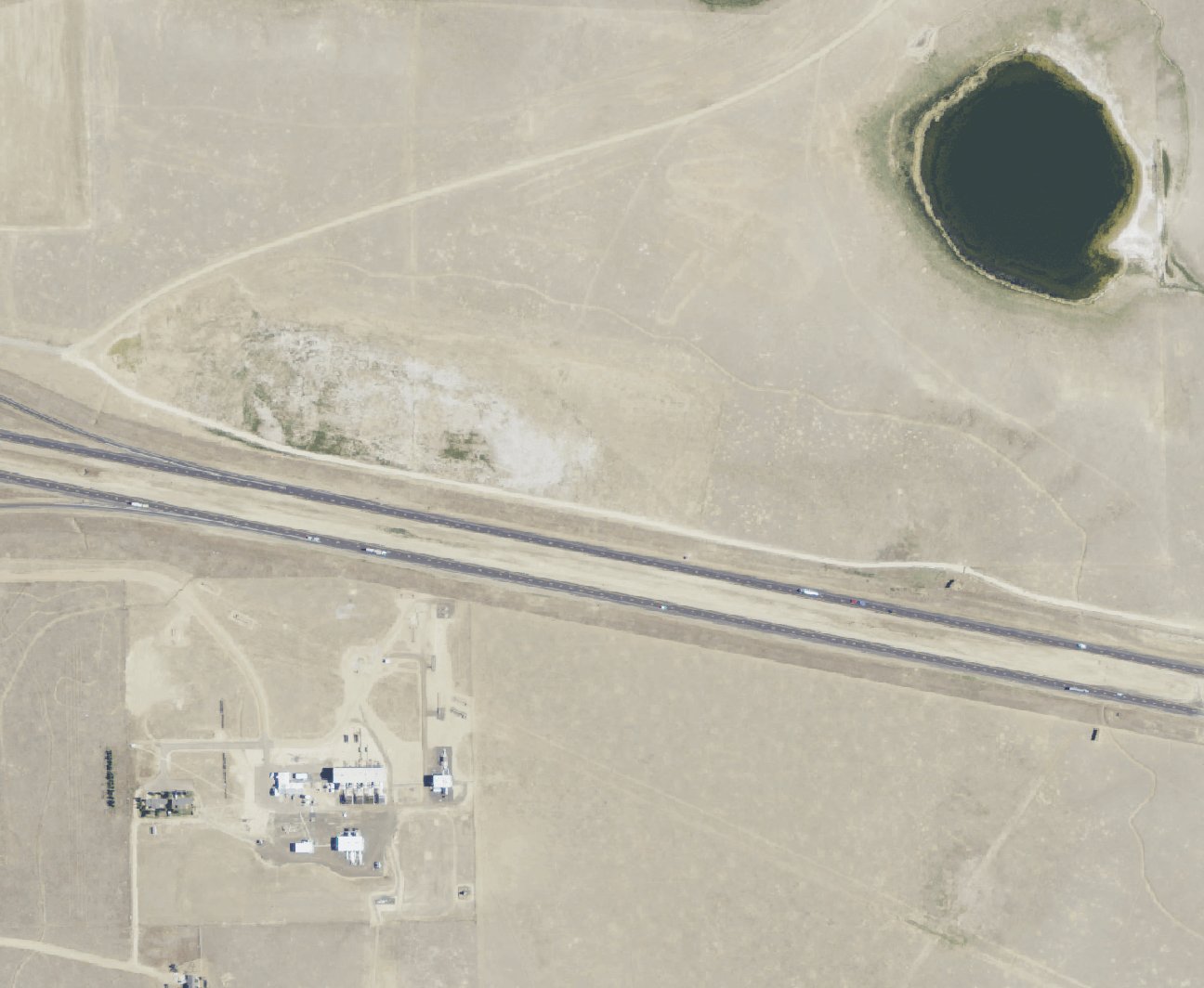}}
		\centerline{(a)}\medskip
	\end{minipage}
	\begin{minipage}[b]{0.49\linewidth}
		\centering
		\centerline{\includegraphics[width=1\linewidth]{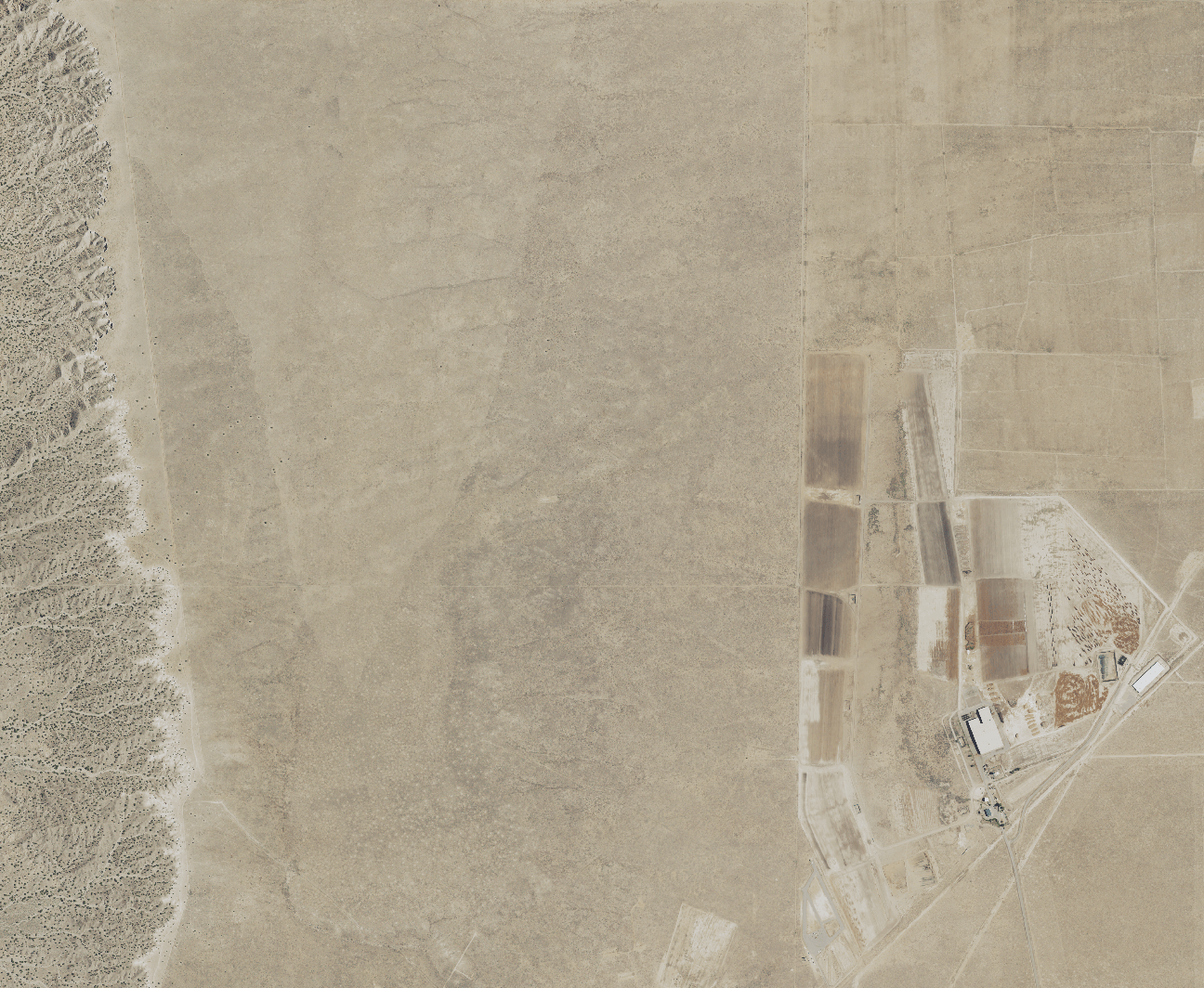}}
		\centerline{(d)}\medskip
	\end{minipage}
	\hfill
	\begin{minipage}[b]{0.49\linewidth}
		\centering
		\centerline{\includegraphics[width=1\linewidth]{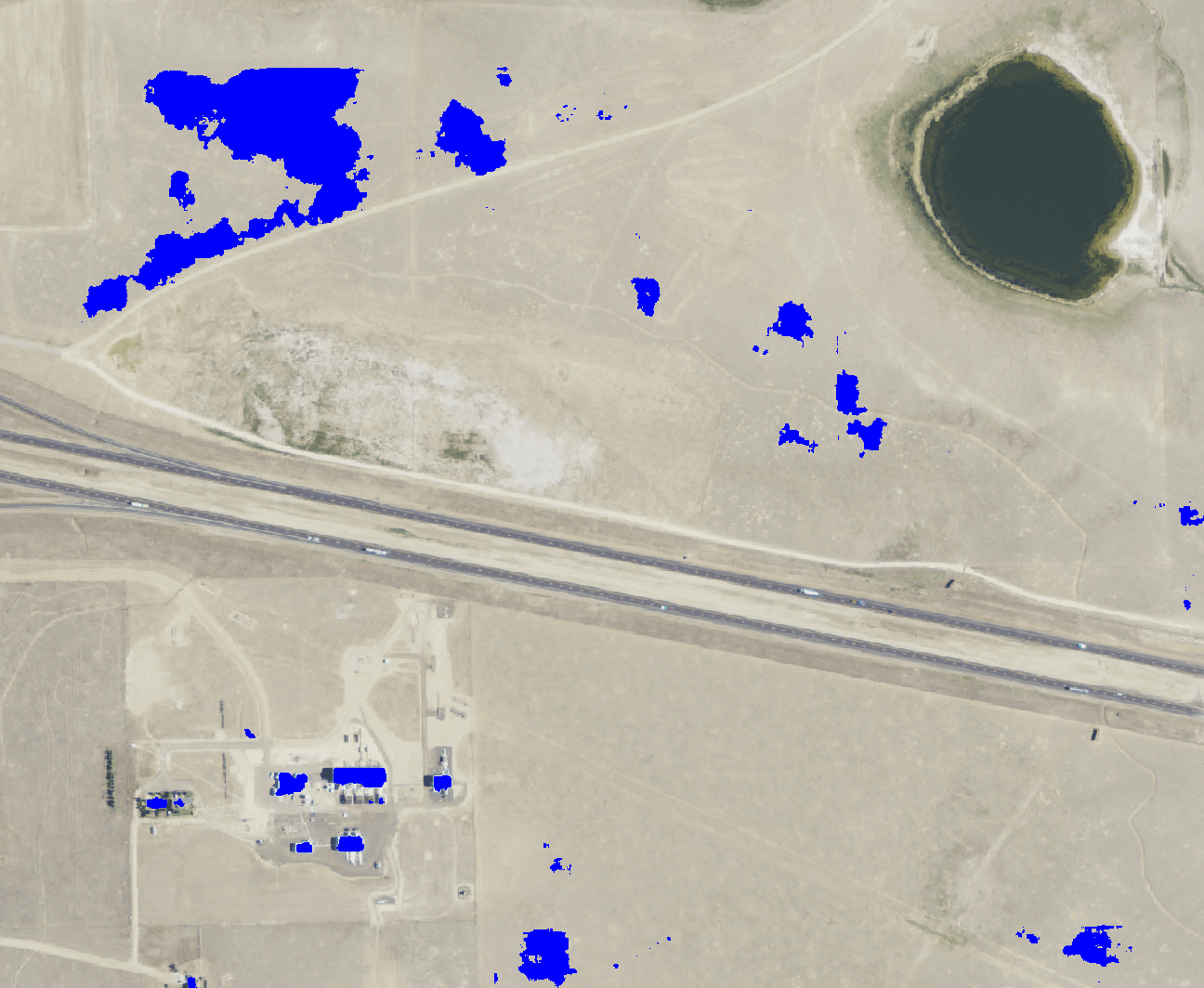}}
		\centerline{(b)}\medskip
	\end{minipage}
	\begin{minipage}[b]{0.49\linewidth}
		\centering
		\centerline{\includegraphics[width=1\linewidth]{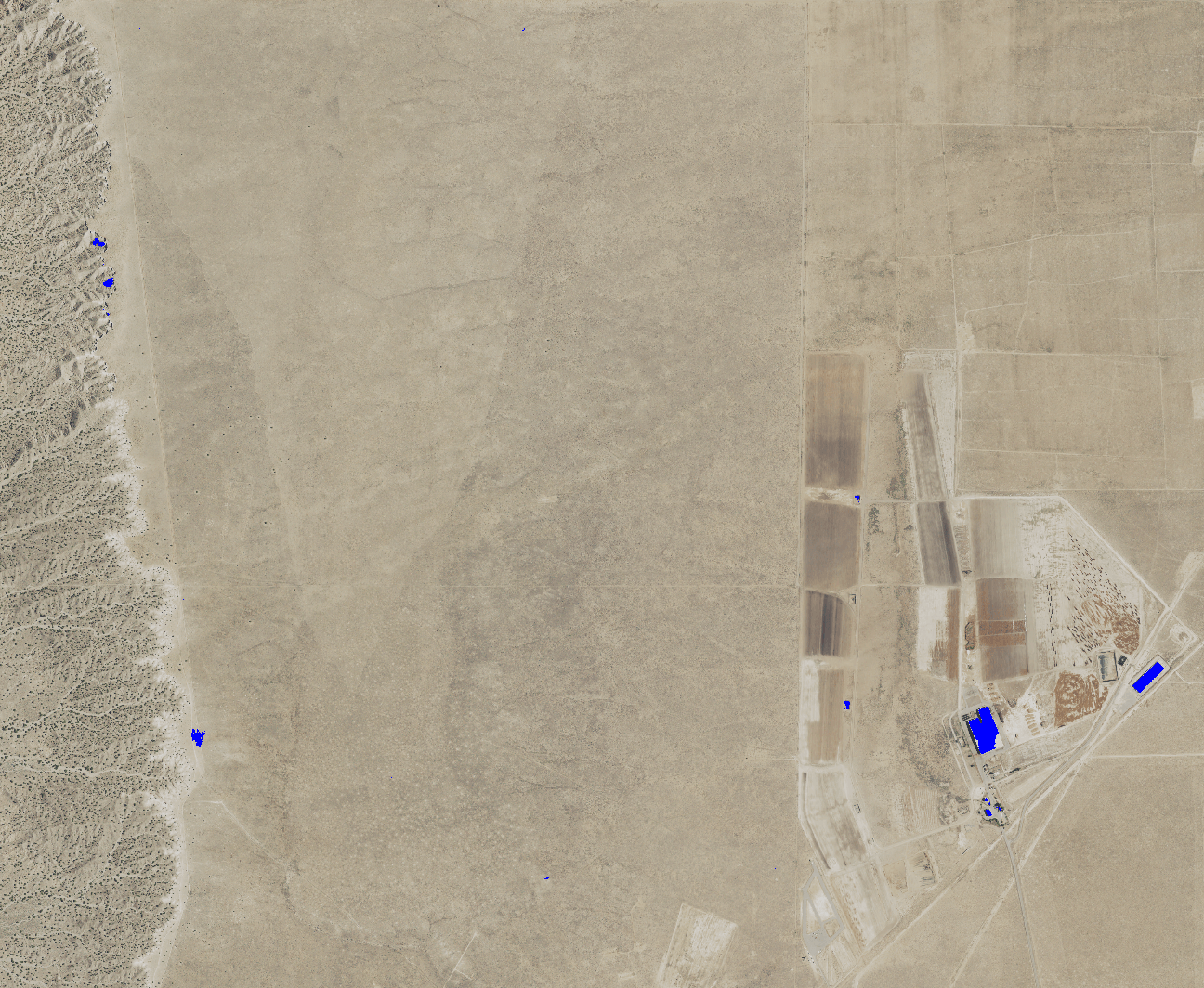}}
		\centerline{(e) }\medskip
	\end{minipage}
\begin{minipage}[b]{0.49\linewidth}
	\centering
	\centerline{\includegraphics[width=1\linewidth]{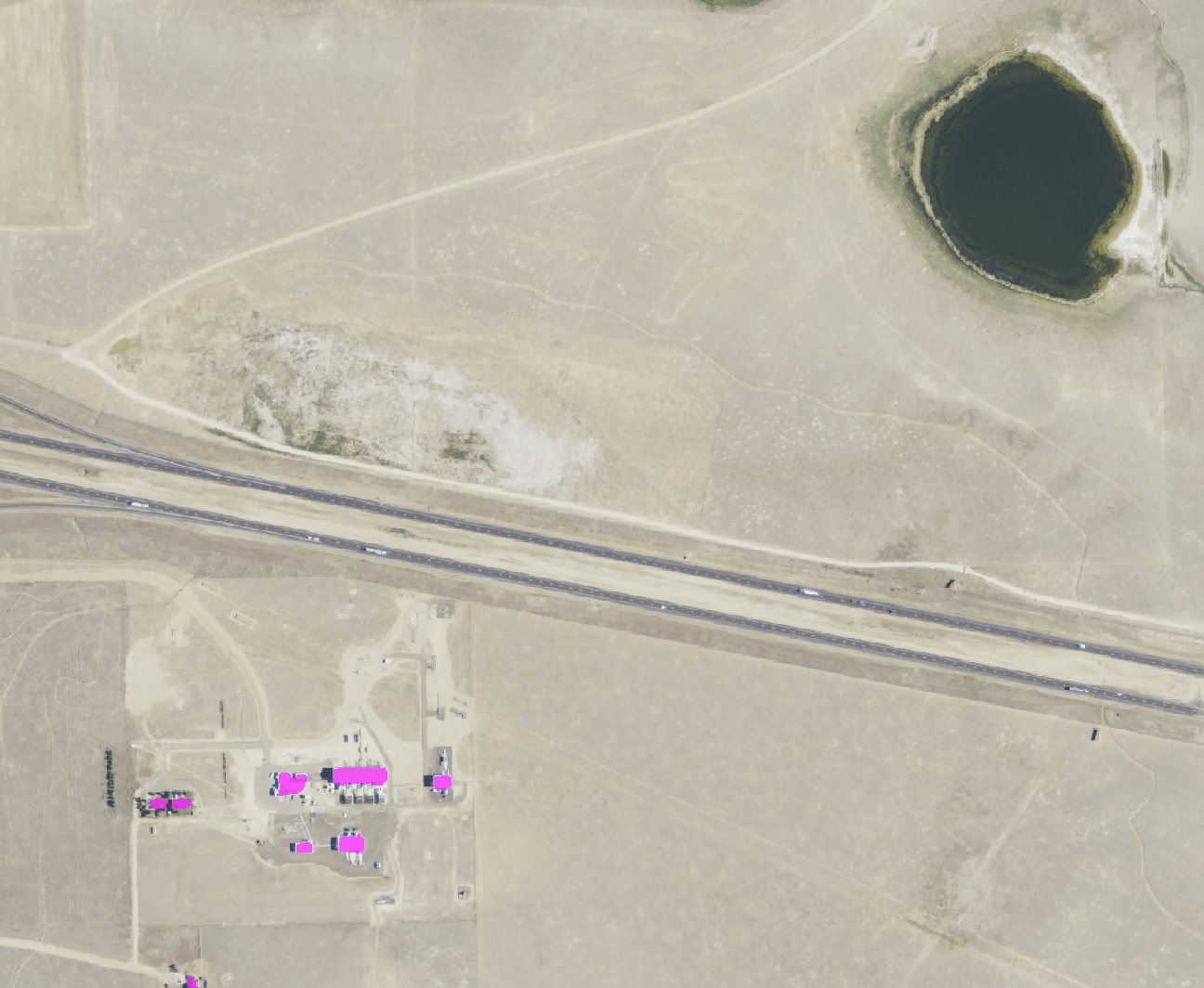}}
	\centerline{(c) }\medskip
\end{minipage}
\begin{minipage}[b]{0.49\linewidth}
	\centering
	\centerline{\includegraphics[width=1\linewidth]{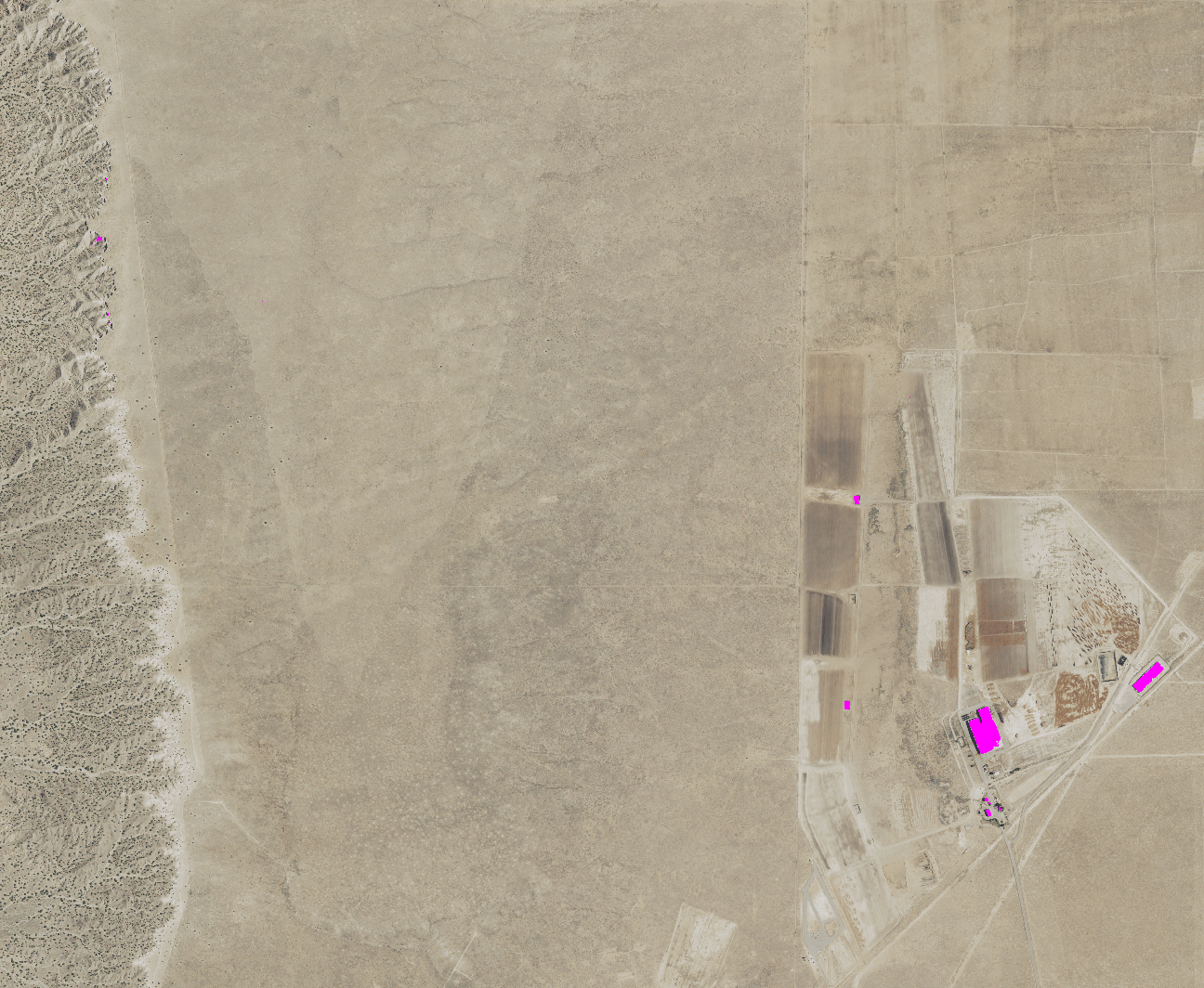}}
	\centerline{(f)}\medskip
\end{minipage}
	\caption{\label{fig:wy_ex1_retrain}Examples of the re-trained model: The images on the left column are from Wyoming and those shown in the right column are from New Mexico. (a), (d) are the input RGB images, (b), (e) are the results (in blue) obtained with the previous model, and (e) and (f) are the results (in purple) provided by the re-trained model.}

\end{figure}

We calculated band statistics of all of the NAIP images ($\sim$ 220,000 images) at including mean, max and min. At state level, we found that the mean values of Wyoming, in particular, is an outlier along with other states that with similar landscape such as New Mexico, Nevada, Arizona, Utah and Colorado. The deployed model is still capable of extracting building from these states, however, more false positives are observed in the desert and mountainous area. To refine the building extraction results, we simply incorporated 141 additional training samples, including 108 negative training samples (no buildings in these samples), and 13 positive training samples (few buildings are manually labeled) from Wyoming. Then, the model was retrained with the original training samples and these additional training samples. After retraining, we tested the newly trained model on the Wyoming images. An example of the results from the original model and the re-trained model are shown in Figure \ref{fig:wy_ex1_retrain} (a), (b) and (c). As we can see, the re-trained model (c) effectively cleans the false positives (b) given by the original model without sacrificing the performance of the building extraction. In addition, with the retrained model we further re-processed the images of other states that exhibit the similar issue, and found out that by only including a small amount of the additional training samples from Wyoming, the retrained model also greatly reduced the false positives of those states whose landscape and mean values of R, G, and B bands are similar to Wyoming. One of the examples from New Mexico is illustrated in Figure \ref{fig:wy_ex1_retrain} (d), (e), (f) where similar observations can be made as in Figure \ref{fig:wy_ex1_retrain} (a), (b), (c).

As noted, the inconsistent image quality and the largely varying terrain types across extensive areas poses major challenges requiring further work to improve CNN-based building extraction. These challenges can be further investigated under transfer learning in the context of deep learning \cite{Yosinski2014}. Future investigations on efficient domain adaptation with CNNs and on the strategy of selecting representative training samples \cite{GeYu2017} might shed some lights on further advancing the power of using CNNs for large scale building extraction. 

\section{Conclusion}
There are two major challenges recognized among many deep learning-based classification tasks as well as in building extractions: 1) require large amount of training samples and 2) need longer training time and demand computational resources. To establish accurate building maps with CNN for the United States, in this paper we addressed the first challenge with diverse training samples collected from different geographical areas and additional negative training samples. The pre-trained models were also used to reduce training time to address the second challenge. 

In order to obtain reliable building extraction results, we evaluated four state-of-the-art semantic segmentation approaches and proposed utilizing max-pooled indices in combination with signed-distance labels to enable  
accurate building extraction results at instance level. Moreover, we proposed a simple but effective fusing strategy that successfully boost the performance of building extraction results. We have established the unprecedented high resolution and seamless building maps for the contiguous United States with the proposed architecture and further refined building extraction results by identifying the major sources of commission errors. The generated model output covering the entire US goes through manual quality checks for verification of the model output. The impressive results have so far been informative for decision making at scale during Irma and Harvey hurricanes. The work continues to benefit other large scale object detection works based on remote sensing imagery.

In the future, we will investigate the benefits of exploiting high performance computing (HPC) resources for multi-gpu training to further advance the fusing strategy. The exploration of HPC will also enable testing more sophisticated but complicated network architectures. It will potentially benefit building extraction or in general object detection with remote sensing imagery.  Currently, the limitations such as the smaller batch size and the size of CNN architectures, are imposed by the capacity of GPU memory and latency between GPU nodes. In addition, with larger GPU memory capacity, we could also investigate the performance of using more than three spectral bands in one CNN for building extraction.
The topic regarding how to select representative samples will also be studied in the future to promote effective domain adaptation in CNNs.


%


\ifCLASSOPTIONcaptionsoff
  \newpage
\fi



%
\bibliographystyle{IEEEbib}

\bibliography{refs_hly}

\begin{thebibliography}{10}

\bibitem{JensenCowen1999}
Jr~Jensen and Dc~Cowen,
\newblock ``{Remote sensing of urban suburban infrastructure and socio-economic
  attributes},''
\newblock {\em Photogrammetric Engineering and Remote Sensing}, vol. 65, no. 5,
  pp. 611--622, 1999.

\bibitem{Ok2013a}
Ali~Ozgun Ok,
\newblock ``{Automated detection of buildings from single VHR multispectral
  images using shadow information and graph cuts},''
\newblock {\em ISPRS Journal of Photogrammetry and Remote Sensing}, vol. 86,
  pp. 21--40, 2013.

\bibitem{Ngo2017}
Tran-thanh Ngo, Vincent Mazet, Christophe Collet, and Paul~De Fraipont,
\newblock ``{Shape-Based Building Detection in Visible Band Images Using Shadow
  Information},''
\newblock {\em IEEE Journal of Selected Topics in Applied Earth Observations
  and Remote Sensing}, vol. 10, no. 3, pp. 920--932, 2017.

\bibitem{Li2015}
Er~Li, John Femiani, Shibiao Xu, Xiaopeng Zhang, and Peter Wonka,
\newblock ``{Robust rooftop extraction from visible band images using higher
  order CRF},''
\newblock {\em IEEE Transactions on Geoscience and Remote Sensing}, vol. 53,
  no. 8, pp. 4483--4495, 2015.

\bibitem{Kim1999}
Taejung Kim and Jan-peter Muller,
\newblock ``{Development of a graph-based approach for building detection},''
\newblock {\em Image and Vision Computing}, vol. 17, no. 1, pp. 3--14, 1999.

\bibitem{Hermosilla2011}
Txomin Hermosilla, Luis~A. Ruiz, Jorge~A. Recio, and Javier Estornell,
\newblock ``{Evaluation of automatic building detection approaches combining
  high resolution images and LiDAR data},''
\newblock {\em Remote Sensing}, vol. 3, no. 6, pp. 1188--1210, 2011.

\bibitem{HaklayWeber2008}
Mordechai~(Muki) Haklay and Patrick Weber,
\newblock ``Openstreetmap: User-generated street maps,''
\newblock {\em IEEE Pervasive Computing}, vol. 7, no. 4, pp. 12--18, Oct. 2008.

\bibitem{PaisitkriangkraiSherrahJanneyEtAl2016}
Sakrapee Paisitkriangkrai, Jamie Sherrah, Pranam Janney, and Anton van~den
  Hengel,
\newblock ``Semantic labeling of aerial and satellite imagery,''
\newblock {\em IEEE Journal of Selected Topics in Applied Earth Observations
  and Remote Sensing}, vol. 9, no. 7, pp. 2868--2881, 2016.

\bibitem{Yuan2018}
J.~Yuan,
\newblock ``Learning building extraction in aerial scenes with convolutional
  networks,''
\newblock {\em IEEE Transactions on Pattern Analysis and Machine Intelligence},
  vol. PP, no. 99, pp. 1--1, 2018.

\bibitem{Bittner2017481}
K.~Bittner, S.~Cui, and P.~Reinartz,
\newblock ``Building extraction from remote sensing data using fully
  convolutional networks,''
\newblock {\em International Archives of the Photogrammetry, Remote Sensing and
  Spatial Information Sciences - ISPRS Archives}, vol. 42, no. 1W1, pp.
  481--486, 2017.

\bibitem{7729471}
Z.~Huang, G.~Cheng, H.~Wang, H.~Li, L.~Shi, and C.~Pan,
\newblock ``Building extraction from multi-source remote sensing images via
  deep deconvolution neural networks,''
\newblock in {\em 2016 IEEE International Geoscience and Remote Sensing
  Symposium (IGARSS)}, July 2016, pp. 1835--1838.

\bibitem{Lunga2018}
D.~Lunga, H.~L. Yang, A.~Reith, J.~Weaver, J.~Yuan, and B.~Bhaduri,
\newblock ``Domain-adapted convolutional networks for satellite image
  classification: A large-scale interactive learning workflow,''
\newblock {\em IEEE Journal of Selected Topics in Applied Earth Observations
  and Remote Sensing}, vol. PP, no. 99, pp. 1--16, 2018.

\bibitem{Jensen1999}
Jr~Jensen and Dc~Cowen,
\newblock ``{Remote sensing of urban suburban infrastructure and socio-economic
  attributes},''
\newblock {\em Photogrammetric Engineering and Remote Sensing}, vol. 65, no. 5,
  pp. 611--622, 1999.

\bibitem{Cote2013}
Melissa Cote and Parvaneh Saeedi,
\newblock ``{Automatic rooftop extraction in nadir aerial imagery of suburban
  regions using corners and variational level set evolution},''
\newblock {\em IEEE Transactions on Geoscience and Remote Sensing}, vol. 51,
  no. 1, pp. 313--328, 2013.

\bibitem{Goodfellow2016}
Ian Goodfellow, Yoshua Bengio, and Aaron Courville,
\newblock {\em Deep Learning},
\newblock MIT Press, 2016,
\newblock \url{http://www.deeplearningbook.org}.

\bibitem{LeCunBottouBengioEtAl1998}
Yann LeCun, L{\'{e}}on Bottou, Yoshua Bengio, and Patrick Haffner,
\newblock ``{Gradient-based learning applied to document recognition},''
\newblock {\em Proceedings of the IEEE}, vol. 86, no. 11, pp. 2278--2323, 1998.

\bibitem{SimonyanZisserman2015}
Karen Simonyan and Andrew Zisserman,
\newblock ``{Very deep convolutional networks for large-scale image
  recognition},''
\newblock in {\em International Conference in Learning Representation}, 2015,
  pp. 1--14.

\bibitem{SzegedyLiuJiaEtAl2015}
C.~Szegedy, Wei Liu, Yangqing Jia, P.~Sermanet, S.~Reed, D.~Anguelov, D.~Erhan,
  V.~Vanhoucke, and A.~Rabinovich,
\newblock ``Going deeper with convolutions,''
\newblock in {\em 2015 IEEE Conference on Computer Vision and Pattern
  Recognition (CVPR)}, June 2015, pp. 1--9.

\bibitem{KarpathyTodericiShettyEtAl2014}
Andrej Karpathy, George Toderici, Sanketh Shetty, Thomas Leung, Rahul
  Sukthankar, and Li~Fei-Fei,
\newblock ``Large-scale video classification with convolutional neural
  networks,''
\newblock in {\em Proceedings of the IEEE conference on Computer Vision and
  Pattern Recognition}, 2014, pp. 1725--1732.

\bibitem{OquabBottouLaptevEtAl2014}
Maxime Oquab, Leon Bottou, Ivan Laptev, and Josef Sivic,
\newblock ``Learning and transferring mid-level image representations using
  convolutional neural networks,''
\newblock in {\em Proceedings of the IEEE conference on computer vision and
  pattern recognition}, 2014, pp. 1717--1724.

\bibitem{SainathKingsburySaonEtAl2015}
Tara~N Sainath, Brian Kingsbury, George Saon, Hagen Soltau, Abdel-rahman
  Mohamed, George Dahl, and Bhuvana Ramabhadran,
\newblock ``Deep convolutional neural networks for large-scale speech tasks,''
\newblock {\em Neural Networks}, vol. 64, pp. 39--48, 2015.

\bibitem{Simonyan2015}
Karen Simonyan and Andrew Zisserman,
\newblock ``{Very deep convolutional networks for large-scale image
  recognition},''
\newblock {\em Iclr}, pp. 1--14, 2015.

\bibitem{ShelhamerLongDarrell2017}
E.~Shelhamer, J.~Long, and T.~Darrell,
\newblock ``Fully convolutional networks for semantic segmentation,''
\newblock {\em IEEE Transactions on Pattern Analysis and Machine Intelligence},
  vol. 39, no. 4, pp. 640--651, April 2017.

\bibitem{NoordPostma2017}
Nanne van Noord and Eric Postma,
\newblock ``Learning scale-variant and scale-invariant features for deep image
  classification,''
\newblock {\em Pattern Recognition}, vol. 61, pp. 583 -- 592, 2017.

\bibitem{Sherrah2016}
Jamie Sherrah,
\newblock ``{Fully Convolutional Networks for Dense Semantic Labelling of
  High-Resolution Aerial Imagery},''
\newblock {\em arXiv}, pp. 1--22, 2016.

\bibitem{Maggiori2016}
Emmanuel Maggiori, Yuliya Tarabalka, Guillaume Charpiat, and Pierre Alliez,
\newblock ``{Fully convolutional neural networks for remote sensing image
  classification},''
\newblock {\em IEEE Transactions on Geoscience and Remote Sensing}, pp.
  5071--5074, 2016.

\bibitem{VolpiTuia2017}
Michele Volpi and Devis Tuia,
\newblock ``{Dense semantic labeling of subdecimeter resolution images with
  convolutional neural networks},''
\newblock {\em IEEE Transactions on Geoscience and Remote Sensing}, vol. 55,
  no. 2, pp. 881--893, 2017.

\bibitem{ErhanManzagolBengioEtAl2009}
Dumitru Erhan, Pierre-Antoine Manzagol, Yoshua Bengio, Samy Bengio, and Pascal
  Vincent,
\newblock ``{The difficulty of training deep architectures and the effect of
  unsupervised pre-training},''
\newblock in {\em Twelfth International Conference on Artificial Intelligence
  and Statistics (AISTATS)}, 2009, vol.~5, pp. 153--160.

\bibitem{Penatti2015}
Otavio A~B Penatti, Keiller Nogueira, and Jefersson~A. {Dos Santos},
\newblock ``{Do deep features generalize from everyday objects to remote
  sensing and aerial scenes domains?},''
\newblock {\em IEEE Computer Society Conference on Computer Vision and Pattern
  Recognition Workshops}, vol. 2015-October, pp. 44--51, 2015.

\bibitem{TajbakhshShinGuruduEtAl2016}
N.~Tajbakhsh, J.~Y. Shin, S.~R. Gurudu, R.~T. Hurst, C.~B. Kendall, M.~B.
  Gotway, and J.~Liang,
\newblock ``Convolutional neural networks for medical image analysis: Full
  training or fine tuning?,''
\newblock {\em IEEE Transactions on Medical Imaging}, vol. 35, no. 5, pp.
  1299--1312, May 2016.

\bibitem{Gerrand2017}
Jonathan Gerrand, Quentin Williams, Dalton Lunga, Adam Pantanowitz, Shabir
  Madhi, and Nasreen Mahomed,
\newblock ``Paediatric frontal chest radiograph screening with fine-tuned
  convolutional neural networks,''
\newblock in {\em Medical Image Understanding and Analysis}, Mar{\'i}a
  Vald{\'e}s~Hern{\'a}ndez and V{\'i}ctor Gonz{\'a}lez-Castro, Eds., Cham,
  2017, pp. 850--861, Springer International Publishing.

\bibitem{ChengMaZhouEtAl2016}
G.~Cheng, C.~Ma, P.~Zhou, X.~Yao, and J.~Han,
\newblock ``Scene classification of high resolution remote sensing images using
  convolutional neural networks,''
\newblock in {\em 2016 IEEE International Geoscience and Remote Sensing
  Symposium (IGARSS)}, July 2016, pp. 767--770.

\bibitem{RadfordMetzChintala2015}
Alec Radford, Luke Metz, and Soumith Chintala,
\newblock ``{Unsupervised Representation Learning with Deep Convolutional
  Generative Adversarial Networks},''
\newblock pp. 1--15, 2015.

\bibitem{ZhengJayasumanaRomera-ParedesEtAl2015}
Shuai Zheng, Sadeep Jayasumana, Bernardino Romera-Paredes, Vibhav Vineet,
  Zhizhong Su, Dalong Du, Chang Huang, and Philip~HS Torr,
\newblock ``Conditional random fields as recurrent neural networks,''
\newblock in {\em Proceedings of the IEEE International Conference on Computer
  Vision}, 2015, pp. 1529--1537.

\bibitem{BadrinarayananKendallCipolla2017}
Alex~Kendall Vijay~Badrinarayanan and Roberto Cipolla,
\newblock ``Segnet: A deep convolutional encoder-decoder architecture for image
  segmentation,''
\newblock {\em IEEE Transactions on Pattern Analysis and Machine Intelligence},
  2017.

\bibitem{silberman2014instance}
Nathan Silberman, David Sontag, and Rob Fergus,
\newblock ``Instance segmentation of indoor scenes using a coverage loss,''
\newblock in {\em European Conference on Computer Vision}. Springer, 2014, pp.
  616--631.

\bibitem{WangBaiMattyusEtAl2016}
Shenlong Wang, Min Bai, Gellert Mattyus, Hang Chu, Wenjie Luo, Bin Yang, Justin
  Liang, Joel Cheverie, Sanja Fidler, and Raquel Urtasun,
\newblock ``{TorontoCity: Seeing the World with a Million Eyes},''
\newblock {\em arXiv preprint}, 2016.

\bibitem{Mnih2013}
Volodymyr Mnih,
\newblock {\em Machine learning for aerial image labeling},
\newblock Ph.D. thesis, University of Toronto (Canada), 2013.

\bibitem{JinDavis2005}
Xiaoying Jin and Curt~H. Davis,
\newblock ``{Automated Building Extraction from High-Resolution Satellite
  Imagery in Urban Areas Using Structural, Contextual, and Spectral
  Information},''
\newblock {\em EURASIP Journal on Advances in Signal Processing}, vol. 2005,
  no. 14, pp. 2196--2206, 2005.

\bibitem{VakalopoulouKarantzalosKomodakisEtAl2015}
M.~Vakalopoulou, K.~Karantzalos, N.~Komodakis, and N.~Paragios,
\newblock ``{Building detection in very high resolution multispectral data with
  deep learning features},''
\newblock {\em 2015 IEEE International Geoscience and Remote Sensing Symposium
  (IGARSS)}, pp. 1873--1876, 2015.

\bibitem{Quionero-CandelaSugiyamaSchwaighoferEtAl2009}
Joaquin Quionero-Candela, Masashi Sugiyama, Anton Schwaighofer, and Neil~D.
  Lawrence,
\newblock {\em Dataset Shift in Machine Learning},
\newblock The MIT Press, 2009.

\bibitem{YuanCheriyadat2014}
Jiangye Yuan and Anil~M Cheriyadat,
\newblock ``Learning to count buildings in diverse aerial scenes,''
\newblock in {\em Proceedings of the 22nd ACM SIGSPATIAL International
  Conference on Advances in Geographic Information Systems}. ACM, 2014, pp.
  271--280.

\bibitem{MaggioriTarabalkaCharpiatEtAl2016}
Emmanuel Maggiori, Yuliya Tarabalka, Guillaume Charpiat, and Pierre Alliez,
\newblock ``{Fully convolutional neural networks for remote sensing image
  classification},''
\newblock {\em IEEE Transactions on Geoscience and Remote Sensing}, pp.
  5071--5074, 2016.

\bibitem{XieTu2015}
Saining Xie and Zhuowen Tu,
\newblock ``{Holistically-Nested Edge Detection},''
\newblock {\em 2015 IEEE International Conference on Computer Vision (ICCV)},
  pp. 1395--1403, 2015.

\bibitem{Ok2013}
Ali~Ozgun Ok, Caglar Senaras, and Baris Yuksel,
\newblock ``{Automated detection of arbitrarily shaped buildings in complex
  environments from monocular VHR optical satellite imagery},''
\newblock {\em IEEE Transactions on Geoscience and Remote Sensing}, vol. 51,
  no. 3, pp. 1701--1717, 2013.

\bibitem{Yosinski2014}
Jason Yosinski, Jeff Clune, Yoshua Bengio, and Hod Lipson,
\newblock ``{How transferable are features in deep neural networks?},''
\newblock {\em Advances in Neural Information Processing Systems 27
  (Proceedings of NIPS)}, vol. 27, pp. 1--9, 2014.

\bibitem{GeYu2017}
Weifeng Ge and Yizhou Yu,
\newblock ``{Borrowing treasures from the wealthy: deep transfer learning
  through selective joint fine-tuning},''
\newblock {\em 2017 IEEE Conference on Computer Vision and Pattern Recognition
  (CVPR)}, pp. 1086--1095, 2017.

\end{thebibliography}

%

%






\end{document}